\documentclass[lettersize,journal]{IEEEtran}
\usepackage{amsmath,amsfonts}
\usepackage{algorithmic}
\usepackage{algorithm}
\usepackage{array}
\usepackage[caption=false,font=normalsize,labelfont=sf,textfont=sf]{subfig}
\usepackage{textcomp}
\usepackage{stfloats}
\usepackage{url}
\usepackage{verbatim}
\usepackage{graphicx}
\usepackage{cite}
\usepackage{booktabs}
\usepackage{multirow}
\usepackage[table]{xcolor}
\usepackage[colorlinks, linkcolor=blue, anchorcolor=blue, citecolor=blue]{hyperref}
\usepackage{pifont}

\begin{document}

\title{STAGE: Tackling Semantic Drift in Multimodal Federated Graph Learning}

\author{Zekai Chen, Xun Wu, Xunkai Li, Yihan Sun, Rong-Hua Li, Guoren Wang
\thanks{Zekai Chen, Xun Wu, Xunkai Li, Rong-Hua Li, Guoren Wang are with Beijing Institute of Technology, Beijing, 100081, China.(e-mail:zackchen02@163.com;alicewu0624@gmail.com;cs.xunkai.li@gmail.com;
lironghuabit@126.com; wanggrbit@gmail.com) Yihan Sun is with Minzu University of China, Beijing, 100081, China.(e-mail:13520392817@163.com)}
}

% The paper headers
\markboth{Journal of \LaTeX\ Class Files,~Vol.~XX, No.~X, May~2026}%
{Anonymous \MakeLowercase{\textit{et al.}}: STAGE: A Differentiable Protocol for Multimodal Federated Graph Learning}

\maketitle

\begin{abstract}
Federated graph learning (FGL) enables collaborative training on graph data across multiple clients.
As graph data increasingly contain multimodal node attributes such as text and images, multimodal federated graph learning (MM-FGL) has become an important yet substantially harder setting.
The key challenge is that clients from different modality domains may not share a common semantic space: even for the same concept, their local encoders can produce inconsistent representations before collaboration begins.
This makes direct parameter coordination unreliable and further causes two downstream problems: forcing heterogeneous client representations into a naively shared semantic space may create false semantic agreement, and graph message passing may amplify residual inconsistency across neighborhoods.
To address this issue, we propose \textbf{STAGE}, a protocol-first framework for MM-FGL.
Instead of relying on direct parameter averaging, STAGE builds a shared semantic space that first translates heterogeneous multimodal features into comparable representations and then regulates how these representations propagate over local graph structures.
In this way, STAGE not only improves cross-client semantic calibration, but also reduces the risk of inconsistency amplification during graph learning.
Extensive experiments on 8 multimodal-attributed graphs across 5 graph-centric and modality-centric tasks show that STAGE consistently achieves state-of-the-art performance while reducing per-round communication payload.
\end{abstract}


\section{Introduction}
\label{sec:intro}

\IEEEPARstart{F}{ederated} graph learning (FGL) enables collaborative training on graph data across multiple clients while preserving privacy by avoiding the sharing of raw data~\cite{he2021fedgraphnn,wu2025comprehensive}. As graph data increasingly carry rich modalities such as text and images, multimodal FGL (MM-FGL) has become an important yet substantially harder setting~\cite{li2026mm}. Crucially, the underlying challenge is not merely statistical Non-IID data distribution. Rather, MM-FGL weakens a fundamental assumption behind standard FGL: isolated clients may no longer produce comparable representations for the same semantic concept, thereby severely disrupting cross-client knowledge aggregation.

\begin{figure}[!t]
    \centering
    \includegraphics[width=\linewidth]{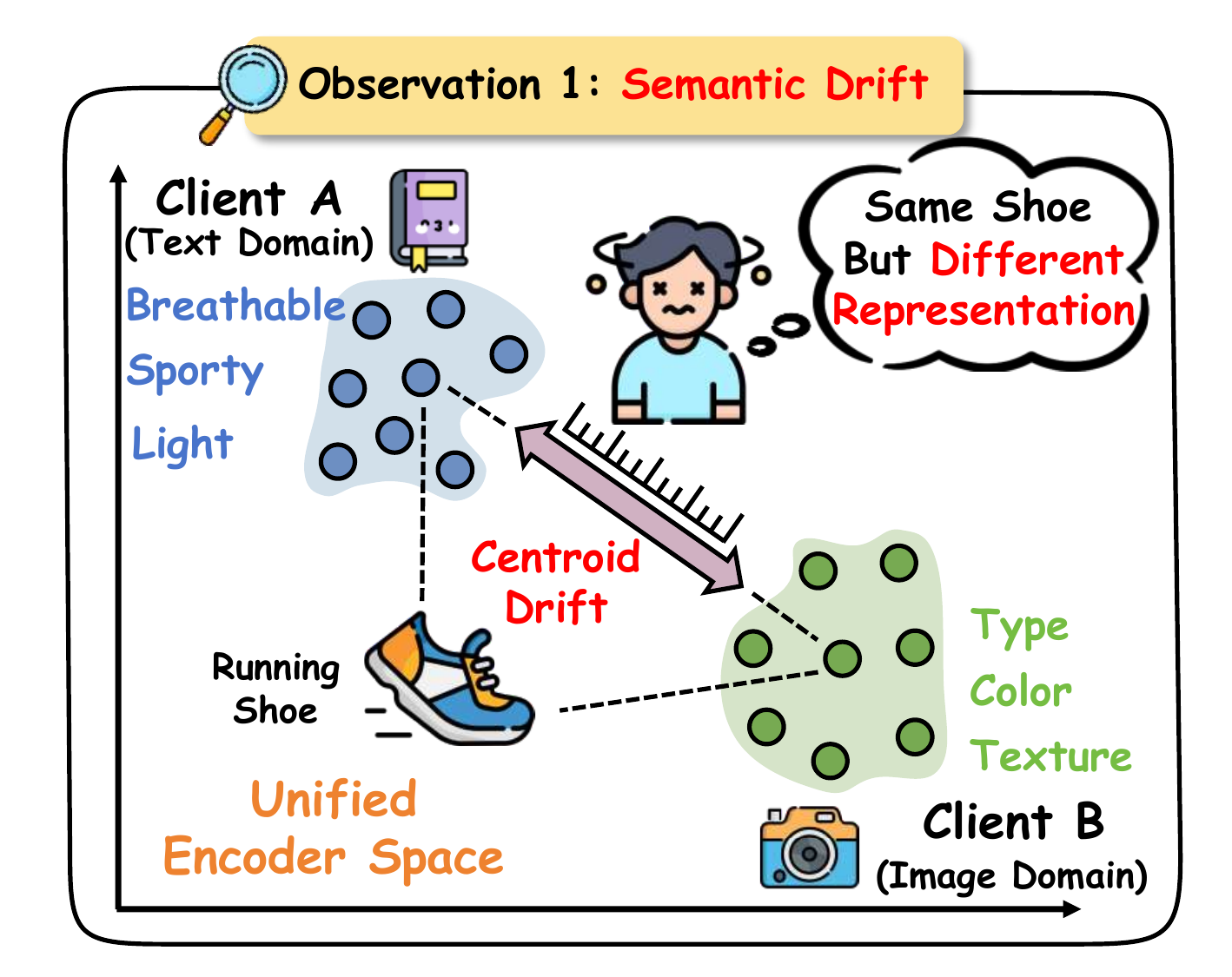}
    \caption{
    \textbf{Observation 1: Semantic drift.}
    Even for the same semantic concept, clients from different modality domains can produce clearly separated representations, leading to large centroid drift in a shared diagnostic space.
    }
    \label{fig:motivation1}
\end{figure}

This challenge points to a more fundamental issue than conventional client drift. In MM-FGL, modality-heterogeneous encoders can map the same concept into different feature regions before collaboration even begins. We refer to this root failure as \textbf{feature drift}. Unlike ordinary feature skew, feature drift reflects a breakdown of \emph{cross-client semantic comparability}: clients may refer to the same concept, yet their learned representations are not directly aligned. Consider the category \emph{running shoes}. A text-dominant client may emphasize words such as \emph{breathable}, \emph{sporty}, and \emph{light}, whereas an image-dominant client may rely more on \emph{type}, \emph{color}, and \emph{texture}. Although both clients refer to the same class, their embeddings can still be far apart. To make this phenomenon measurable, we map node representations into a shared frozen diagnostic space and compute the distance between class centroids across clients. As shown in Fig.~\ref{fig:motivation1}, the centroid drift of the same class is already large under modality skew, indicating that MM-FGL may lack semantic comparability before meaningful collaboration starts. In Sec.~\ref{sec:empirical}, we further revisit this observation through layer-wise visualization and quantitative drift analysis (Fig.~\ref{fig:emp_semantic_drift} and Fig.~\ref{fig:emp_propagation_bar}).

This observation also explains why existing paradigms remain fundamentally insufficient. Classical federated optimization improves robustness under statistical non-IID data distributions, but it still assumes that clients can be coordinated through a common parameter space~\cite{mcmahan2017communication,li2020federated,karimireddy2020scaffold}. Personalized and heterogeneous Federated Learning (FL) methods relax this rigid assumption by exchanging prototypes, logits, or compact representation statistics~\cite{tan2022fedproto,zhang2024fedtgp,yu2023multimodal}. However, these methods still rely on an implicit prerequisite that proves exceptionally fragile in MM-FGL: clients must already produce roughly comparable representations. Once this foundational prerequisite fails, improving server-side aggregation alone is inherently insufficient to recover true semantic consistency across clients.

\begin{figure}[!t]
    \centering
    \includegraphics[width=\linewidth]{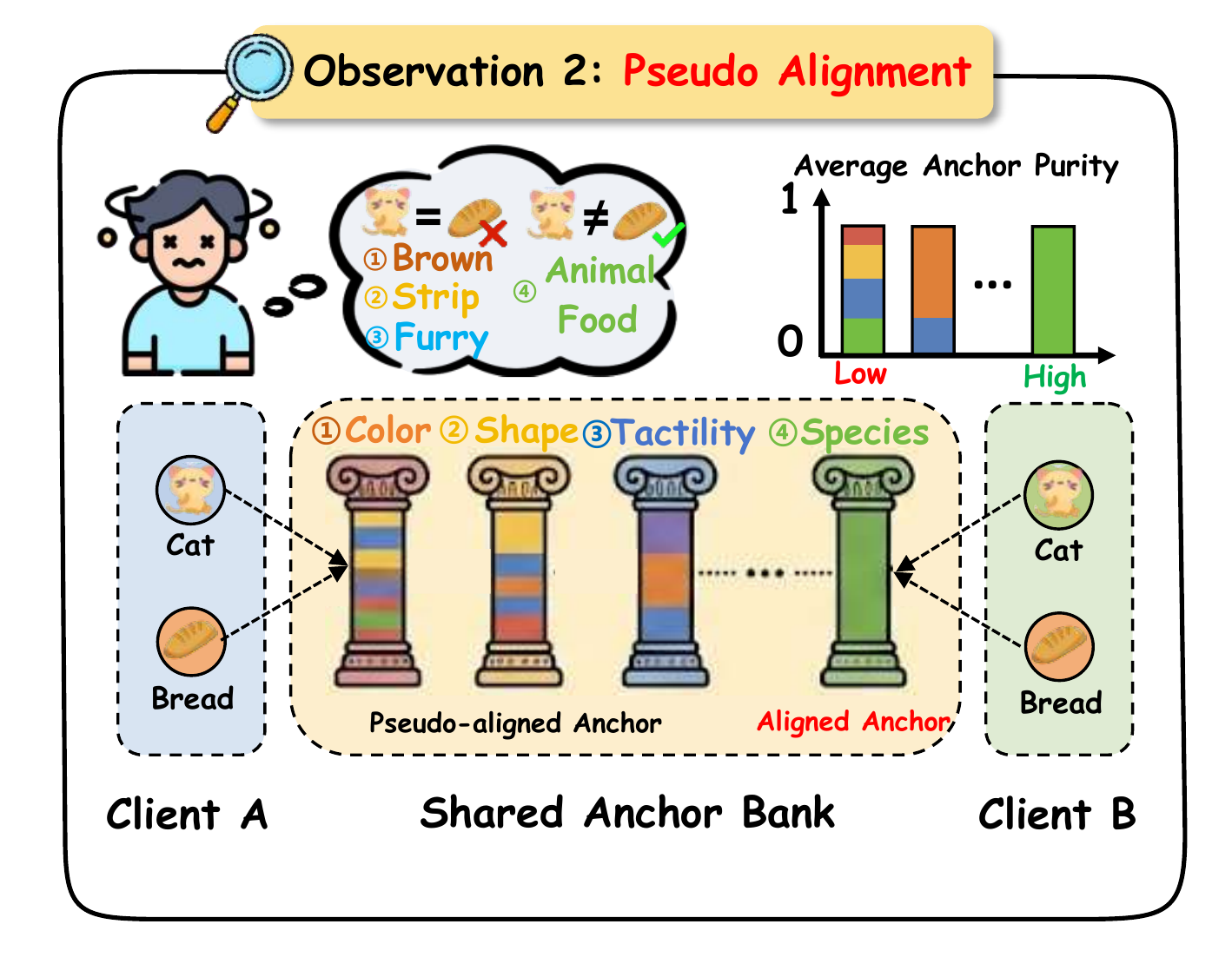}
    \caption{
    \textbf{Observation 2: Pseudo-alignment.}
    A naively shared anchor bank does not guarantee shared semantics: different clients may map different meanings to the same anchor.
    }
    \label{fig:motivation2}
\end{figure}

A seemingly natural remedy is to introduce a shared semantic space across clients. Yet in MM-FGL, naive anchor sharing is not sufficient. If semantically incompatible local features are projected into a shared anchor space without global calibration, different clients may still map different meanings to the same anchor. We term this secondary failure \emph{pseudo-alignment}. To quantify it, we measure \emph{anchor purity}: if an anchor corresponds to stable semantics, it should be dominated by one coherent concept; if it mixes unrelated concepts, its meaning is unstable. Fig.~\ref{fig:motivation2} shows that anchor purity remains low without global semantic calibration, indicating that sharing anchors is not the same as sharing semantics. In Sec.~\ref{sec:empirical}, we analyze this failure more directly through dominant-class purity gains and anchor-class assignment deltas (Fig.~\ref{fig:dominant_class_purity}).

More importantly, pseudo-alignment is not the end of the story. In graph learning, residual semantic inconsistency does not remain isolated to individual nodes. Once inconsistent node semantics enter message passing, iterative graph propagation amplifies this mismatch into a cascading downstream failure, termed propagation-induced drift. As demonstrated in Sec.~\ref{sec:empirical}, this detrimental effect is reflected both in the growth of cross-client centroid drift after graph propagation (Fig.~\ref{fig:emp_propagation_bar}) and in the systematic miscalibration of local attention mechanisms under multimodal skew (Fig.~\ref{fig:emp_attention}). Consequently, the central question in MM-FGL is not simply how to average models more effectively, but how to establish rigorous semantic comparability across clients before parameter aggregation and structural propagation begin. This naturally suggests a protocol-first view of MM-FGL: clients should first communicate through a compact semantic space that proactively reduces feature drift, and only then regulate how aligned semantics propagate over local graph structures.

Motivated by this perspective, we propose \textbf{STAGE} (\textbf{S}emantic \textbf{T}ranslation, \textbf{A}nchor calibration, \textbf{G}raph regulation, and \textbf{E}ntropy regularization), a protocol-first framework for MM-FGL. STAGE separates the problem into two ordered stages. First, it introduces \textbf{Variational Semantic Calibration}, which maps multimodal node features into low-dimensional distributions over a shared frozen anchor bank. To make this space stable across clients, the server maintains \textbf{Global Anchor Prototypes (GAP)} that calibrate local anchor meanings and correct pseudo-alignment. Second, on top of this calibrated representation space, STAGE performs \textbf{Differentiable Homophily Control}, which learns client-specific propagation priors and suppresses the amplification of residual inconsistency during graph message passing. In this way, STAGE explicitly separates \emph{making semantics comparable} from \emph{controlling how those semantics propagate}.

\textbf{Contributions.}
\underline{\textbf{\textit{(1) New Challenge.}}}
We identify \textbf{feature drift} as the root obstacle in MM-FGL---the breakdown of cross-client semantic comparability before collaboration---and empirically expose its downstream consequences through centroid drift and anchor purity analyses (Sec.~\ref{sec:empirical}).
\underline{\textbf{\textit{(2) New Framework.}}}
We propose \textbf{STAGE}, which first reduces feature drift through semantic translation and server-side anchor calibration, and then suppresses its downstream consequences through graph regulation and entropy regularization.
\underline{\textbf{\textit{(3) SOTA Performance.}}}
Across multimodal non-IID benchmarks, STAGE consistently outperforms strong FGL baselines by 4.62\% while reducing communication payload more than 100$\times$ through low-dimensional protocol messages.

\section{Problem Formulation}
\label{sec:problem}

We consider a MM-FGL system of $K$ decentralized clients~\cite{mcmahan2017communication,he2021fedgraphnn}.
Client $k$ holds a private multimodal-attributed graph $\mathcal{G}_k=(\mathcal{V}_k,\mathcal{E}_k)$, where each node $v$ carries a masked multimodal input $\mathbf{x}_v=\{m_v^{(c)}\cdot x_v^{(c)}\}_{c\in\mathcal{M}}$ with availability mask $\mathbf{m}_v\in\{0,1\}^{|\mathcal{M}|}$.
Client $k$ uses a frozen backbone $\psi_k^{(c)}$ with architecture-dependent output dimension $d_c^{(k)}$, and a lightweight trainable projector $\phi_k:\mathbb{R}^{d_k}\to\mathbb{R}^{d_p}$ that maps fused multimodal features into a shared protocol space:
\begin{equation}
\mathbf{h}_v = \phi_k\!\left( \bigoplus_{c\in\mathcal{M}} \big(m_v^{(c)}\cdot \psi_k^{(c)}(x_v^{(c)})\big) \right) \in \mathbb{R}^{d_p}.
\label{eq:fusion_prelim}
\end{equation}
Let $\mathcal{W}_k=\{\phi_k,\Theta_k\}$ denote all trainable parameters. Classical federated learning seeks a consensus model $\bar{\mathcal{W}}$ via
\begin{equation}
\min_{\bar{\mathcal{W}}} \sum_{k=1}^{K} p_k \mathcal{L}_{\mathrm{task}}^{(k)}(\mathcal{W}_k;\mathcal{G}_k)
\quad \text{s.t.} \quad \mathcal{W}_k=\bar{\mathcal{W}},\ \forall k.
\label{eq:naive_fed_prelim}
\end{equation}

\section{Empirical Investigation}
\label{sec:empirical}

Before introducing the full method, we first ask a more fundamental question: \emph{what exactly breaks in multimodal federated graph learning?}
Our central claim is that the core obstacle in MM-FGL is not ordinary statistical Non-IID, but a hierarchy of semantic failures caused by multimodal heterogeneity across clients.
Concretely, we identify a three-stage failure chain:
\ding{182} \textbf{Semantic drift}, where clients do not share a common semantic coordinate system before collaboration begins;
\ding{183} \textbf{Pseudo-alignment}, where a naively shared semantic space creates false consensus rather than true agreement; and
\ding{184} \textbf{Propagation-induced drift}, where residual inconsistency is further amplified by graph message passing.
This section is organized exactly along this hierarchy, which in turn motivates the design of our protocol-first framework.

\paragraph{Stage I: Semantic Drift as the Root Failure.}
\label{sec:stage_1}
Eq.~\eqref{eq:naive_fed_prelim} is not only difficult to optimize, but also conceptually mismatched to MM-FGL.
When $d_c^{(k)} \neq d_c^{(j)}$, the local parameter spaces satisfy $\dim(\Omega_k)\neq\dim(\Omega_j)$, so the consensus constraint $\mathcal{W}_k=\bar{\mathcal{W}}$ is not naturally defined in a shared space.
More importantly, even if some downstream modules can be numerically aligned, parameter agreement does not guarantee semantic comparability.
Under multimodal heterogeneity, the same class may occupy very different feature regions across clients because local encoders induce incompatible semantics before collaboration begins.
We refer to this root failure as \textbf{semantic drift}.

\begin{figure}[htbp]
\centering
\includegraphics[width=\columnwidth]{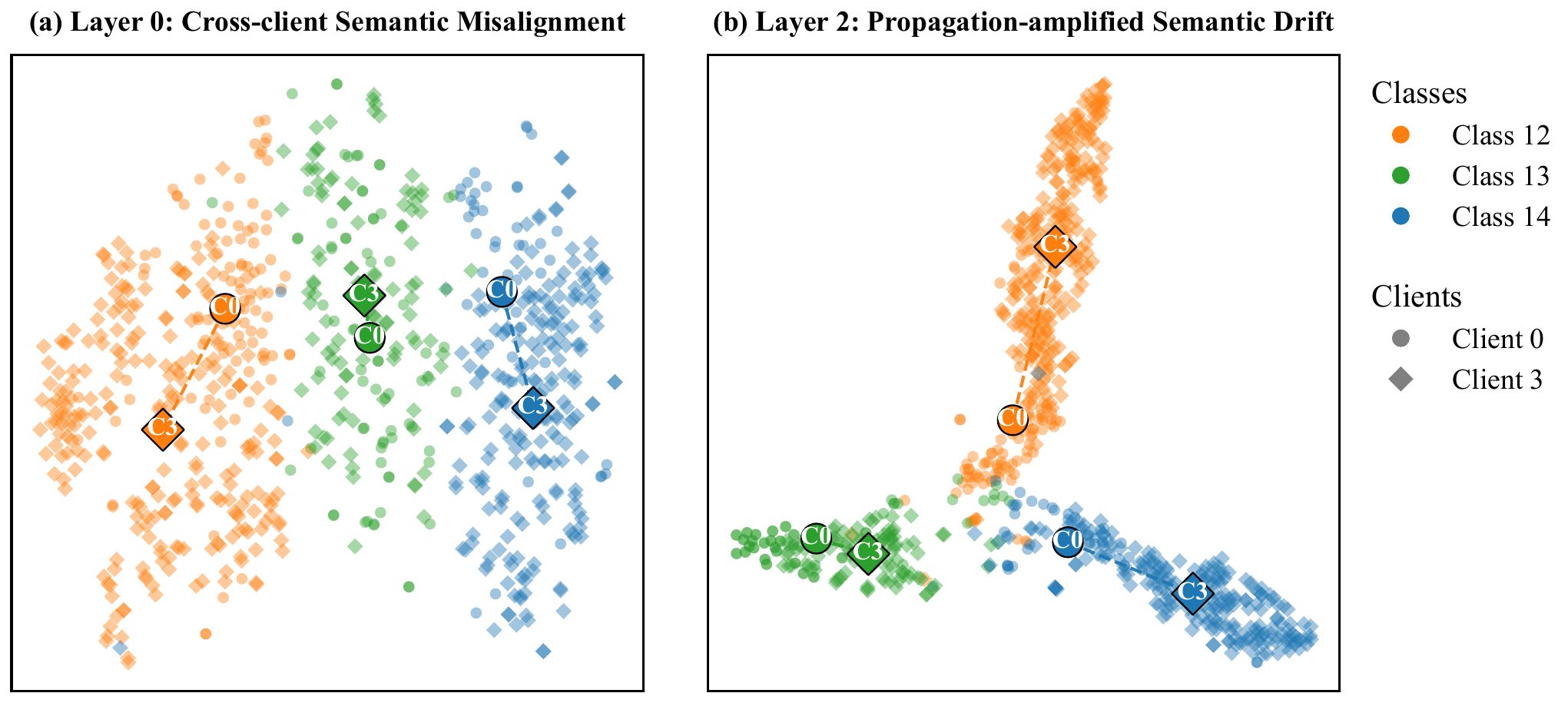}
\caption{\textbf{Empirical observation of the semantic-failure chain on the Toys dataset.}
(Left) At Layer 0, nodes from the same semantic classes are already misaligned across clients, revealing semantic drift before collaboration.
(Right) At Layer 2, this inconsistency becomes more severe after graph propagation, foreshadowing propagation-induced drift.}
\label{fig:emp_semantic_drift}
\end{figure}

To visualize this failure, we project the initial multimodal features (Layer 0) of the Toys dataset into a shared latent space using t-SNE.
As shown in Fig.~\ref{fig:emp_semantic_drift}(Left), nodes from the same class are already clearly separated across clients before meaningful collaboration begins.
For example, the centroids of Class 12 on Client 0 and Client 3 are far apart despite representing the same concept.
This suggests that MM-FGL may fail at the semantic level \emph{before} federated optimization begins.

\begin{figure}[htbp]
\centering
\includegraphics[width=0.8\columnwidth]{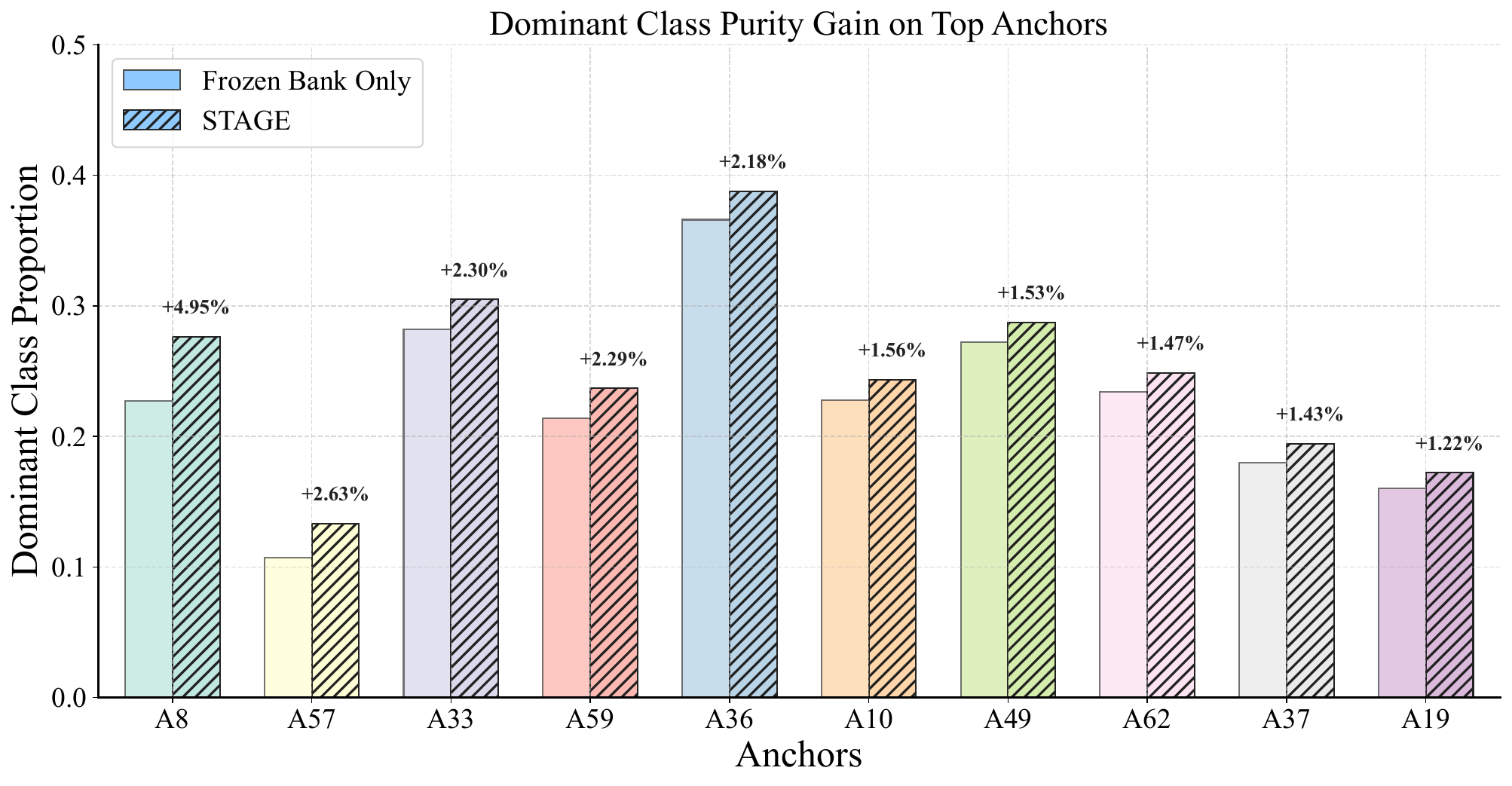}
\caption{\textbf{Validation of pseudo-alignment correction.}
Dominant-class purity gain across top activated anchors.}
\label{fig:dominant_class_purity}
\end{figure}

\paragraph{Stage II: Pseudo-Alignment as the Failure of Naive Shared Anchors.}
A natural response to semantic drift is to impose a shared semantic space, e.g., through a frozen anchor bank.
However, this alone does not ensure true agreement.
If incompatible local features are projected into a shared anchor space without global calibration, different clients may map different meanings to the same anchor, creating a \emph{false consensus}.
We refer to this as \textbf{pseudo-alignment}.

To show how our protocol resolves this issue, we analyze the class distributions of the most frequently activated anchors.
As shown in Fig.~\ref{fig:dominant_class_purity}, STAGE consistently increases the dominant-class proportion across major anchors, indicating that anchor semantics become more concentrated and less ambiguous after calibration.

\begin{figure}[htbp]
\centering
\includegraphics[width=0.8\columnwidth]{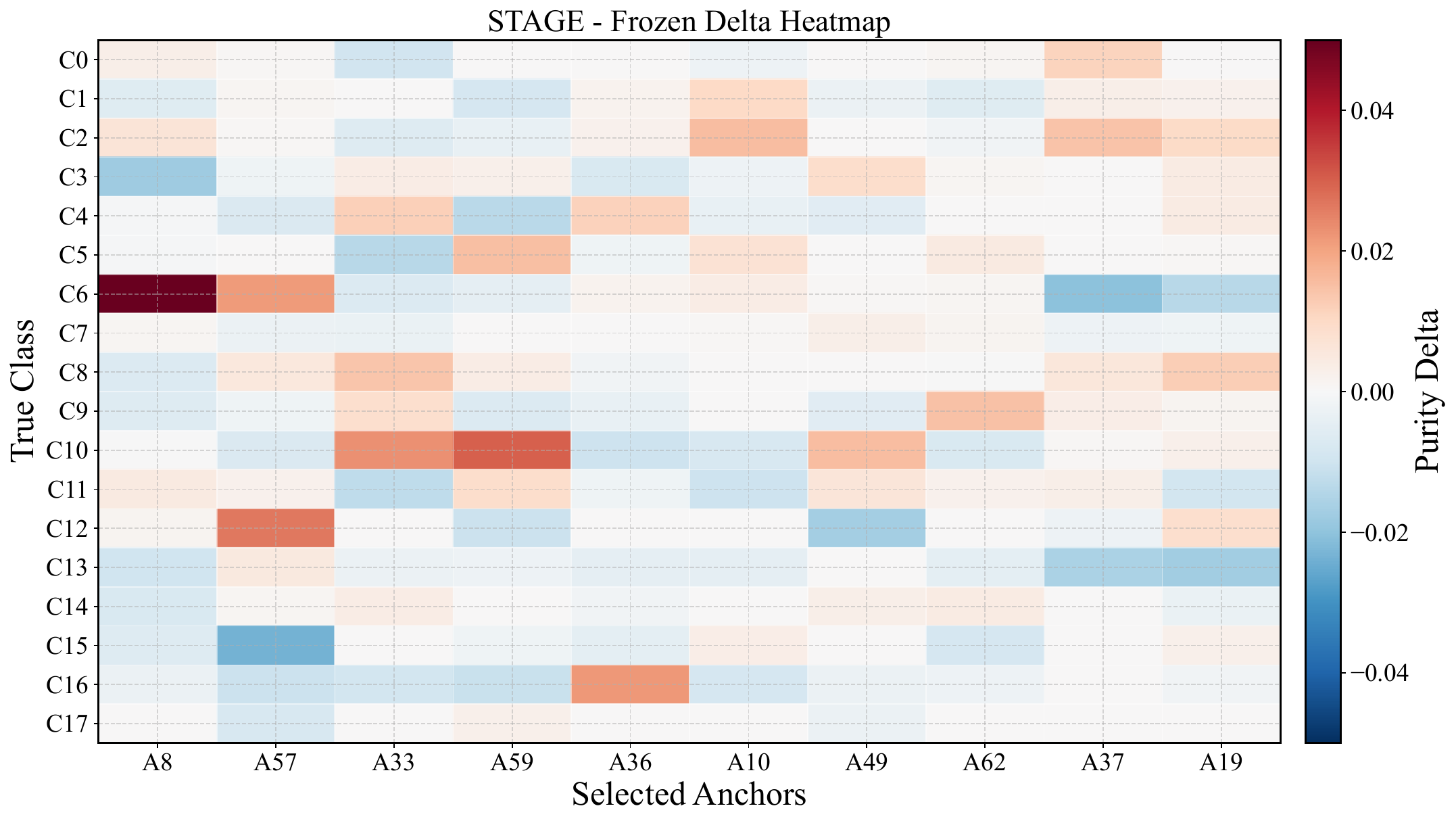}
\caption{\textbf{Validation of pseudo-alignment correction.}
Delta heatmap of anchor-class assignment distributions.}
\label{fig:anchor_delta_heatmap}
\end{figure}

The heatmap in Fig.~\ref{fig:anchor_delta_heatmap} further shows the mechanism:
dominant concepts are reinforced, while heterogeneous assignments are suppressed.
Thus, STAGE does not merely redistribute anchor usage, but purifies the shared semantic vocabulary and reduces false semantic consensus across clients.

\paragraph{Stage III: Propagation-Induced Drift as Downstream Amplification.}
Pseudo-alignment is still not the end of the story. If residual inconsistency remains in the semantic space, graph propagation does not correct it; instead, it can amplify it into a more severe downstream failure, which we term \textbf{propagation-induced drift}. We formalize this effect through the graph \emph{Dirichlet energy}
\begin{equation}
\mathcal{E}(H)=\mathrm{tr}(H^\top L H)=\frac{1}{2}\sum_{(u,v)\in\mathcal{E}} A_{uv}\|\mathbf{h}_u-\mathbf{h}_v\|^2,
\label{eq:dirichlet}
\end{equation}
where $L=D-A$ is the graph Laplacian. Under multimodal heterogeneity, unresolved semantic inconsistency can be repeatedly propagated and amplified across GNN layers.

\begin{figure}[htbp]
\centering
\includegraphics[width=\columnwidth]{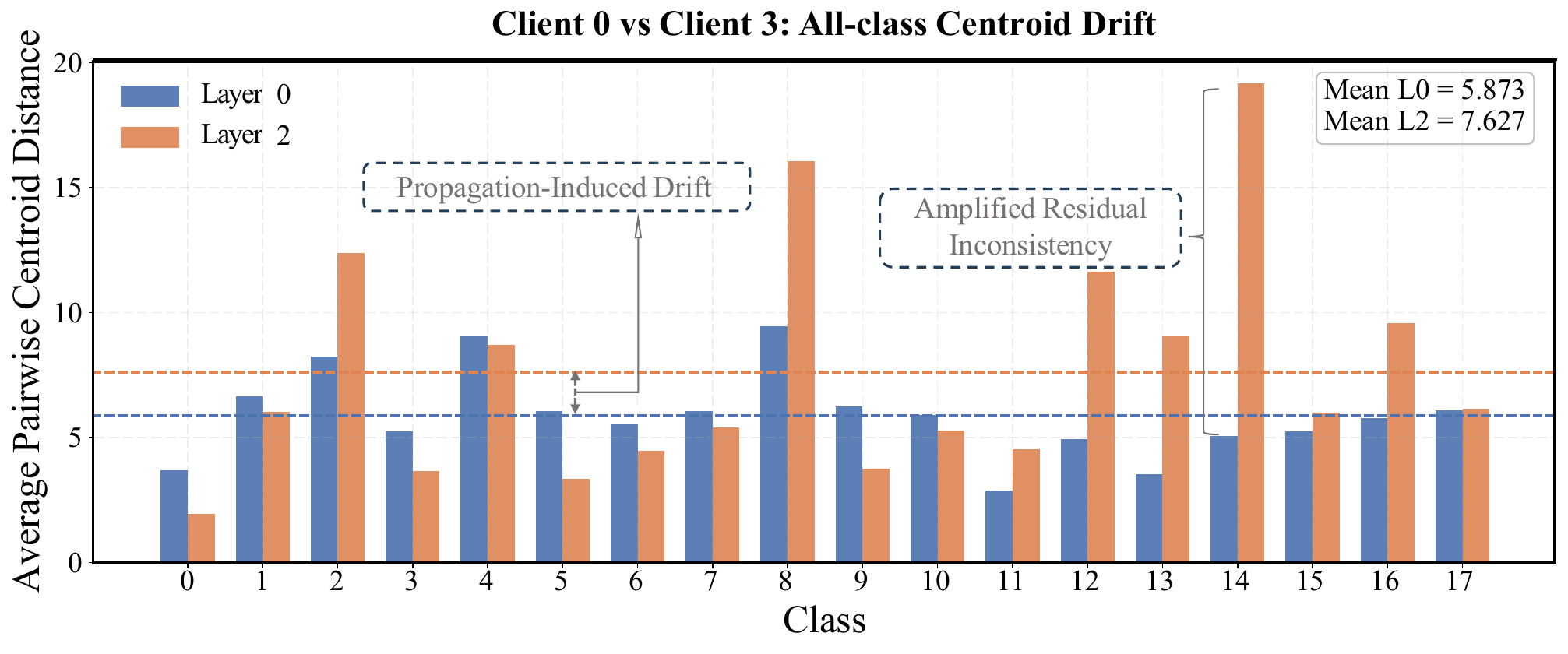}
\caption{\textbf{Quantitative analysis of propagation-induced drift.}
Without semantic calibration, pairwise centroid distances increase substantially after graph propagation, showing that message passing amplifies residual inconsistency rather than correcting it.}
\label{fig:emp_propagation_bar}
\end{figure}

This effect is quantified in Fig.~\ref{fig:emp_propagation_bar}. After two GNN layers, pairwise centroid distances for most classes increase substantially, with the overall mean drift growing by $29.8\%$. This confirms that once semantic inconsistency enters message passing, it becomes a graph-level amplification effect rather than a local representational issue.

A natural question is whether learnable local propagation, such as graph attention, can self-correct this problem. Our answer is no. Since each client only observes a partial structural view, local attention mechanisms cannot reliably infer which neighbors should be trusted under extreme multimodal skew.

\begin{figure}[!t]
    \centering
    \includegraphics[width=\columnwidth]{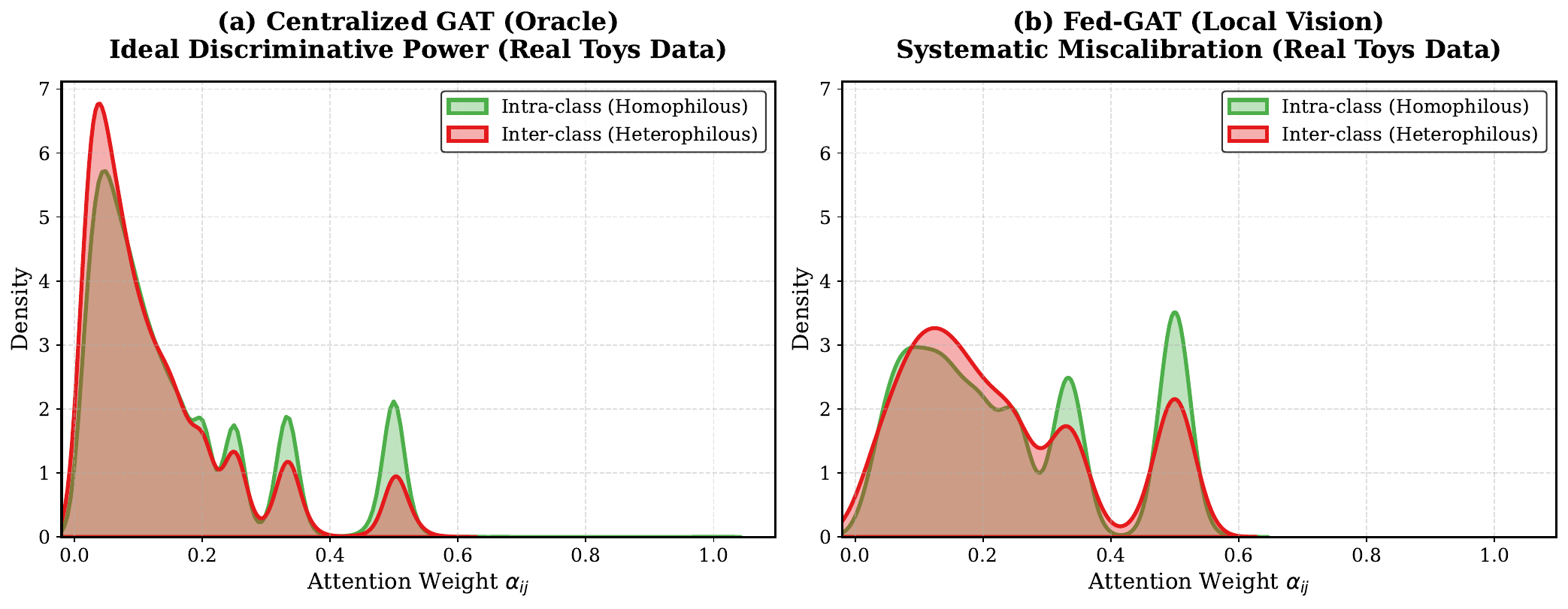}
    \caption{\textbf{Systematic miscalibration of local attention.}
    (Left) A centralized oracle clearly separates homophilous and heterophilous edges.
    (Right) In the federated setting, limited structural exposure causes severe overlap between the two distributions.}
    \label{fig:emp_attention}
\end{figure}

To verify this point, we analyze the attention weights learned by GAT on the Toys dataset, separating intra-class (homophilous) and inter-class (heterophilous) edges. As shown in Fig.~\ref{fig:emp_attention}(Left), a centralized oracle assigns clearly different attention weights to the two edge types, indicating well-calibrated neighbor reliability. In contrast, Fig.~\ref{fig:emp_attention}(Right) shows severe overlap in the federated setting, where local attention fails to calibrate neighbor trust under multimodal skew. This suggests that propagation-induced drift cannot be resolved by local message-passing heuristics alone, but instead requires a global mechanism to regulate how semantics propagate over local graph structures.

Overall, these findings support a coherent failure hierarchy in MM-FGL: \emph{semantic drift} is the root problem, \emph{pseudo-alignment} is the failure of naive shared anchors, and \emph{propagation-induced drift} is the downstream amplification of residual inconsistency. This directly motivates our design principle: MM-FGL should first establish semantic comparability through a shared protocol and then regulate how those semantics propagate.

\section{Related Work}
\label{sec:related}

Our empirical findings suggest that the main difficulty of MM-FGL lies in establishing semantic comparability across clients before graph propagation. Existing methods mainly improve parameter coordination, representation exchange, or centralized multimodal graph modeling, but none simultaneously resolves semantic inconsistency and its propagation over local graph structures.

\textbf{Parameter-Space FGL.}
FedAvg-style graph extensions improve collaboration under topology and label heterogeneity through gradient clustering, personalized subgraph collaboration, topology-aware aggregation, and homophily-aware message passing~\cite{mcmahan2017communication,xie2021federated,baek2023personalized,li2024fedgta,zhu2024fedtad,zheng2025meta}. Other methods use alternative information carriers, such as prototypes or condensed graph structures, to transfer knowledge beyond direct parameter averaging~\cite{tan2022fedproto,chen2025rethinking}. However, these methods still assume that exchanged information is semantically comparable across clients, which becomes fragile under feature drift.

\textbf{Representation-Space HFL/pFL.}
Heterogeneous and personalized FL methods bypass structural non-isomorphism by exchanging soft predictions, class prototypes, or compact multimodal abstractions~\cite{lin2020ensemble,shen2020federated,tan2022fedproto,zhang2024fedtgp,yu2023multimodal,feng2023fedmultimodal}. Yet they still rely on the existence of comparable representations, an assumption that does not hold in MM-FGL when feature drift is severe.

\textbf{Centralized Multimodal-Attributed Graph Learning.}
Centralized MAG methods have advanced modality-topology fusion from early graph-multimodal architectures to recent decoupled frameworks~\cite{li2019semi,lu2019vilbert,chen2024lion,tang2024higpt,he2025unigraph2}. While these methods provide useful architectural insights, they assume a single learner with direct access to all modalities. Therefore, they do not address the semantic inconsistency and propagation issues introduced by decentralized multimodal heterogeneity.
As summarized in Table~\ref{tab:paradigm_comparison}, STAGE is designed to address all three.

\begin{table}[!t]
\centering
\caption{\textbf{Paradigm comparison across three MM-FGL requirements.} Only STAGE satisfies all three.}
\label{tab:paradigm_comparison}
\vspace{2pt}
\setlength{\tabcolsep}{3pt} % 稍微缩小列与列之间的空白间距
\resizebox{\columnwidth}{!}{ % 强制将整个表格缩放至单栏宽度
\begin{tabular}{lccc}
\toprule
\textbf{Paradigm} & \textbf{Graph-topology-aware} & \textbf{Encoder-agnostic} & \textbf{Decentralized} \\
\midrule
Parameter-Space FGL & \checkmark & \texttimes & \checkmark \\
Representation-Space HFL/pFL & \texttimes & \checkmark & \checkmark \\
Centralized MAG & \checkmark & \checkmark & \texttimes \\
\textbf{STAGE (Ours)} & \textbf{\checkmark} & \textbf{\checkmark} & \textbf{\checkmark} \\
\bottomrule
\end{tabular}
}
\end{table}

\IEEEpubidadjcol

\section{Methodology}
\label{sec:method}
\subsection{Overview}
In this section, we present a detailed description of the STAGE framework. Following our protocol-first perspective, we design the framework around three ordered stagesL: (1) Initialization, which establishes a shared semantic reference space; (2) Semantic Alignment, which translates features into probability distributions and globally calibrates local semantics; and (3) Propagation Control, which regulates message passing to suppress residual inconsistency.

\begin{figure*}[!t]
    \centering
    \includegraphics[width=\linewidth]{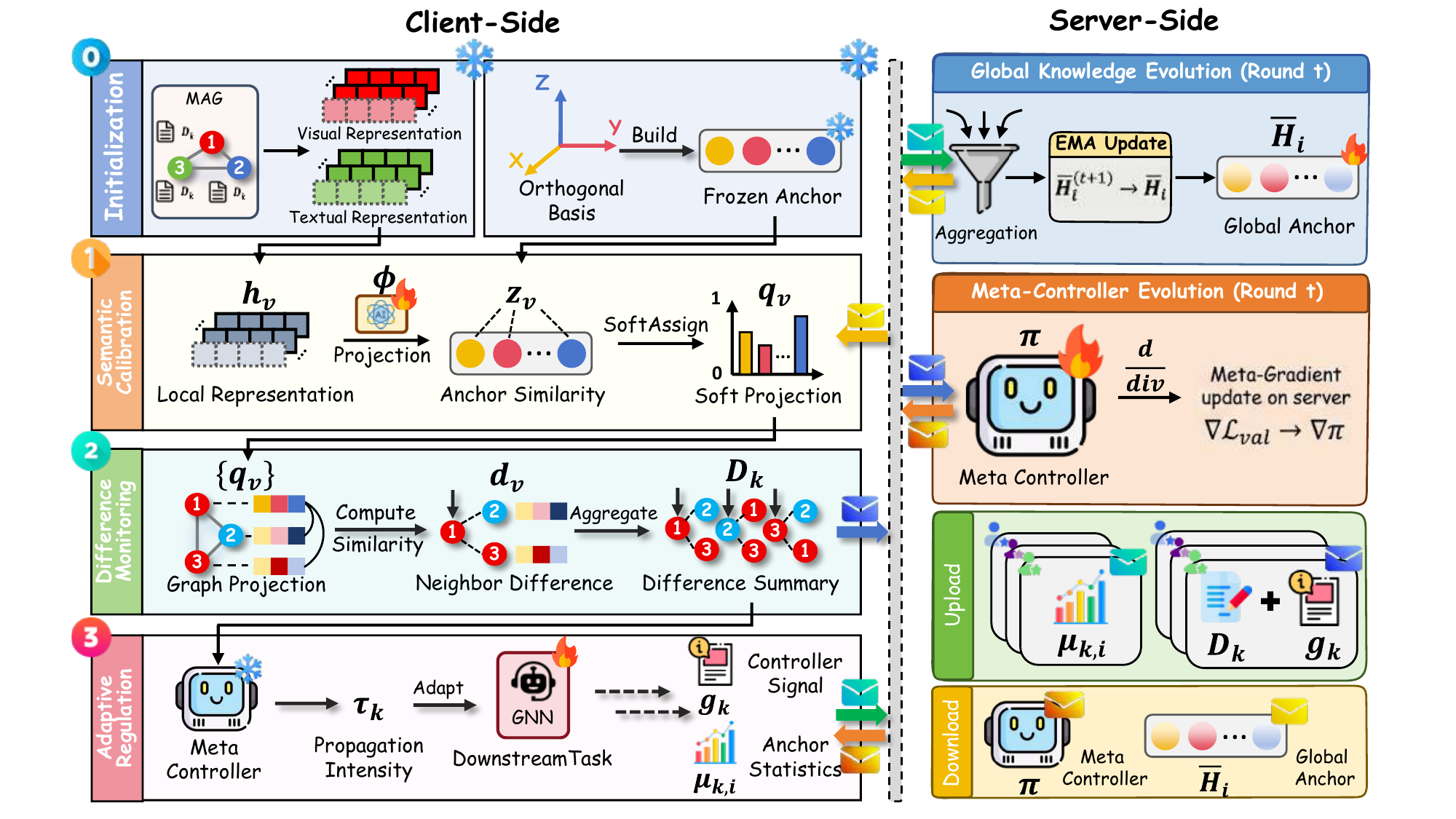}
    \caption{Overview of the STAGE framework. The protocol-first architecture decouples federated graph learning into client execution and server evolution. Local nodes establish a shared reference frame through a frozen semantic bank (Step 0), aligning modality-heterogeneous features via variational soft projection (Step 1). Structural-semantic conflict is summarized into a compact sketch $D_k$ (Step 2), allowing the meta-controller to adaptively regulate propagation intensity $\tau_k$ for downstream GNNs (Step 3). At the server, global consensus is maintained by updating anchor prototypes via exponential moving average and optimizing the meta-controller $\pi$ with uploaded low-dimensional gradients, preserving communication efficiency and privacy.}
    \label{fig:framework}
\end{figure*}

\subsection{Initialization}
\textbf{Motivation.} Our empirical investigation shows that modality heterogeneity induces severe semantic drift, with isolated encoders mapping identical concepts to different feature regions (Fig.~\ref{fig:emp_semantic_drift}). To establish a shared reference before collaboration, we introduce a global semantic bank that bypasses parameter-space misalignment.

\textbf{Multimodal feature extraction.} As shown in the initialization stage of Fig.~\ref{fig:framework}, each client processes a private MAG. For node $v$, the input consists of multiple modalities, such as visual and textual data. We use modality-specific frozen backbones, denoted as $\psi_k^{(c)}$, to process each modality $c \in \mathcal{M}$. Since these isolated encoders project data into heterogeneous spaces with varying dimensions $d_c^{(k)}$, the resulting representations are not directly comparable across clients. This misalignment causes the observed feature drift and motivates a globally calibrated reference.

\textbf{Global anchor bank construction.} To provide a common target for these unaligned multimodal representations, STAGE constructs a globally shared, frozen semantic bank $\mathcal{B} = \{b_i\}_{i=1}^M \in \mathbb{R}^{M \times d_{p}}$. The bank is orthogonally initialized to cover the latent semantic space and remains fixed throughout federated training. Keeping it frozen prevents the reference space from drifting during model aggregation, thereby maintaining stable semantic consistency across clients.

\subsection{Semantic Calibration}
\textbf{Motivation.}
Although the shared bank provides a unified space, isolated local training can still cause pseudo-alignment and protocol collapse (Fig.~\ref{fig:dominant_class_purity}). To ensure consistent anchor semantics without sacrificing representational capacity, we calibrate local translations and maximize entropy.

\textbf{Variational semantic translation.} The first step is to map heterogeneous representations into the shared anchor space. Let $h_v$ denote the multimodal representation of node $v$ on client $k$, and let $\phi_k$ be the local projector. To enforce a consistent statistical manifold across clients, we formulate feature-to-anchor assignment as a Kullback-Leibler (KL) regularized projection onto the probability simplex $\Delta^{M-1}$:
\begin{equation}
\small
\label{eq:kl_projection}
q_v=\arg\min_{q\in\Delta^{M-1}}\left(-\sum_{i=1}^M q^{(i)}\frac{\phi_k(h_v)^\top b_i}{\|\phi_k(h_v)\|_2\|b_i\|_2}+\tau_s D_{KL}(q\parallel u)\right)
\end{equation}
where $u = \frac{1}{M}\mathbf{1}$ is the uniform prior and $\tau_s > 0$ is the entropic temperature. The closed-form solution for this projection is:
\begin{equation}
    \label{eq:translation}
    q_v^{(i)} = \frac{\exp \left( \frac{1}{\tau_s} \frac{\phi_k(h_v)^\top b_i}{\|\phi_k(h_v)\|_2 \|b_i\|_2} \right)}{\sum_{j=1}^M \exp \left( \frac{1}{\tau_s} \frac{\phi_k(h_v)^\top b_j}{\|\phi_k(h_v)\|_2 \|b_j\|_2} \right)}
\end{equation}

This shifts the alignment target from incompatible continuous parameter spaces to a modality-agnostic discrete distribution space, thereby reducing feature drift before propagation.

\textbf{Max-Entropy for information capacity.} However, projecting features into this discrete space may cause mode collapse in early training, where $\phi_k$ maps most nodes to only a few anchors. This reduces the capacity of the frozen bank, distorts anchor statistics, and creates dead anchors. To encourage balanced use of the semantic vocabulary, we apply max-entropy regularization to the global semantic distribution of each client. For client $k$, let $\bar{q}_k = \frac{1}{|\mathcal{V}_k|} \sum_{v \in \mathcal{V}_k} q_v$ denote the average semantic assignment over the local graph. We introduce an entropic penalty that minimizes the negative Shannon entropy of this expected distribution:
\begin{equation}
    \label{eq:entropy}
    \mathcal{L}_{ent}^{(k)} = \sum_{i=1}^M \bar{q}_k^{(i)} \log \bar{q}_k^{(i)}
\end{equation}
Minimizing $\mathcal{L}_{ent}^{(k)}$ encourages local projections to span the discrete space more broadly, improving the discriminative power of the global prototypes.

\textbf{Anchor-conditional statistics and Contrastive GAP.} With representational capacity stabilized, STAGE corrects pseudo-alignment using server-maintained Global Anchor Prototypes, denoted as $\bar{H} \in \mathbb{R}^{M \times d_{p}}$. At each communication round, client $k$ computes the local anchor-conditional mean representation $\mu_{k,i}$ for each active anchor $i$:
\begin{equation}
    \label{eq:anchor_mean}
    \mu_{k,i} = \frac{1}{|\mathcal{V}_{k,i}|} \sum_{v \in \mathcal{V}_{k,i}} \phi_k(h_v)
\end{equation}

where $\mathcal{V}_{k,i} = \{v \in \mathcal{V}_k \mid \arg\max_j q_v^{(j)} = i\}$ is the subset of nodes hard-assigned to anchor $i$. The server aggregates these local statistics to update $\bar{H}$ via an exponential moving average (EMA) to maintain a stable global semantic reference. To enforce semantic consensus across the federation, client $k$ aligns its local anchor semantics with the downloaded GAP using an InfoNCE objective:
\begin{equation}
    \label{eq:gap_loss}
    \mathcal{L}_{gap}^{(k)} = - \frac{1}{|\mathcal{I}_k|} \sum_{i \in \mathcal{I}_k} \log \frac{\exp(sim(\mu_{k,i}, \bar{H}_i) / \tau_c)}{\sum_{j=1}^M \exp(sim(\mu_{k,i}, \bar{H}_j) / \tau_c)}
\end{equation}
where $\mathcal{I}_k$ denotes the set of active anchors on client $k$, $sim(\cdot, \cdot)$ represents the cosine similarity, and $\tau_c$ is the contrastive temperature. This calibration mechanism ensures that the shared discrete simplex reflects a globally consistent semantic vocabulary, effectively mitigating pseudo-alignment.

\subsection{Adaptive Regulation}
\textbf{Motivation.}
Even after semantic calibration, graph propagation may still introduce noise under severe non-IID conditions (Fig.~\ref{fig:emp_attention}). To suppress the amplification of residual inconsistency, we use a Meta-Controller to regulate message passing.

\textbf{Structural-semantic conflict sketch.} To quantify propagation-induced drift, we measure the structural-semantic conflict for each node $v \in \mathcal{V}_k$ by computing the Jensen-Shannon (JS) divergence between its semantic fingerprint $q_v$ and the local neighborhood context $\tilde{q}_v = \sum_{u \in \mathcal{N}(v)} w_{vu} q_u$, where $w_{vu}$ denotes the baseline topological weights. To preserve privacy while capturing local homophily patterns, client $k$ summarizes these node-level discrepancies $d_v = D_{JS}(q_v \parallel \tilde{q}_v)$ into a low-dimensional sketch $D_k \in \mathbb{R}^2$:
\begin{equation}
    \label{eq:sketch}
    D_k = [\mu(d_v), \sigma(d_v)]^\top
\end{equation}
This sketch contains only the mean and standard deviation of local semantic conflicts, avoiding the exposure of raw features or graph topology.

\textbf{Regulated propagation via meta-controller.} Using $D_k$, the server meta-learns a neural controller $\pi_{\theta_\pi}$, which is downloaded to each client to generate a propagation prior $\tau_k = \pi_{\theta_\pi}(D_k)$. We use $\tau_k$ as a temperature factor to adjust local GNN aggregation and suppress heterophilous noise:
\begin{equation}
\label{eq:controlled_alpha}
    \hat{\alpha}_{vu} = \frac{\exp(e_{vu} / \tau_k)}{\sum_{w \in \mathcal{N}(v)} \exp(e_{vw} / \tau_k)}
\end{equation}

\textbf{Communication-efficient meta-update.} To optimize the global controller $\pi_{\theta_\pi}$ without sharing raw topologies or model Hessians, we adopt an implicit bilevel optimization protocol. Client $k$ evaluates the task loss on a local validation set and computes a scalar meta-gradient $g_k = \nabla_{\tau_k} \mathcal{L}_{val}^{(k)} \in \mathbb{R}$. The server then aggregates these low-dimensional signals and updates the controller by the chain rule:
\begin{equation}
    \label{eq:meta_update}
    \theta_\pi \leftarrow \theta_\pi - \eta_\pi \sum_{k=1}^K g_k \cdot \nabla_{\theta_\pi} \pi_{\theta_\pi}(D_k)
\end{equation}
where $\eta_\pi$ is the meta-learning rate. This keeps the communication payload limited to the low-dimensional tuple $(D_k, g_k)$, improving efficiency while preserving privacy.

\section{Experiments}
\label{sec:exp}
In this section, we present a comprehensive evaluation of STAGE. We begin by introducing the experimental setup (Sec.\ref{subsec:exp_setup}), and then seek to answer the following research questions: \textbf{Q1:} Does STAGE consistently outperform SOTA FGL methods under multimodal Non-IID conditions (Sec.\ref{subsec:eq1})? \textbf{Q2:} How does each individual component contribute to the overall performance of STAGE (Sec.\ref{subsec:eq2})? \textbf{Q3:} Does STAGE maintain strong scalability and robustness against client fragmentation and changes in core hyperparameters (Sec.\ref{subsec:eq3})? \textbf{Q4:} Does the protocol-first design substantially reduce communication payload and accelerate training convergence compared to parameter-sharing alternatives (Sec.\ref{subsec:eq4})?

\subsection{Experimental Setup}
\label{subsec:exp_setup}
\textbf{Datasets.} We evaluate STAGE on eight benchmark multimodal-attributed graph datasets covering diverse domains: Toys~\cite{ni2019justifying}, Grocery~\cite{ni2019justifying}, Bili Music~\cite{zhang2024ninerec}, DY~\cite{zhang2024ninerec}, QB~\cite{zhang2024ninerec}, Bili Cartoon~\cite{zhang2024ninerec}, Flickr30k~\cite{plummer2015flickr30k}, and SemArt~\cite{garcia2018read}. To construct federated client partitions, we apply the Louvain algorithm~\cite{blondel2008fast} for node classification, modality retrieval, G2Text, and G2Image. For link prediction, we use METIS~\cite{karypis1998fast} instead, as it minimizes edge cuts and thus better preserves local connectivity for predicting links.

\textbf{Baselines.}
To evaluate whether STAGE addresses the full failure chain in MM-FGL, we compare it with six categories of baselines. 
(1) \textbf{FL}: FedAvg~\cite{mcmahan2017communication} represents generic federated learning based on standard parameter-space consensus. 
(2) \textbf{Multimodal graph collaboration}: Fed-MGNet~\cite{hamilton2017inductive} and Fed-MHGAT~\cite{velivckovic2018graph} serve as multimodal graph baselines that model cross-modal interactions without explicitly regulating decentralized graph propagation. 
(3) \textbf{MM-FL}: FedMVP~\cite{che2024leveraging} and FedMAC~\cite{nguyen2024fedmac} provide comparisons against multimodal federated learning frameworks. 
(4) \textbf{Cross-client discrepancy mitigation}: FedLap~\cite{aliakbari2025subgraph} and S2FGL~\cite{tan2025s2fgl} address client-level inconsistency in federated graph learning, serving as references for methods that reduce cross-client divergence without explicit semantic calibration. 
(5) \textbf{Topology-aware FGL}: FedSPA~\cite{tan2025fedspa} and FedIIH~\cite{yu2025modeling} benchmark federated graph learning methods designed for topology heterogeneity. 
(6) \textbf{Prototype-based transfer}: FedProto~\cite{tan2022fedproto} serves as a representative baseline that transfers cross-client knowledge through compact prototypes rather than direct parameter averaging.through contrastive calibration. 

\textbf{Downstream Tasks.} We evaluate the methods on two groups of downstream tasks. \textbf{Graph-centric tasks} include node classification and link prediction, measured by Accuracy and AUC, respectively. \textbf{Modality-centric tasks} include modality retrieval, G2Text, and G2Image, measured by Recall@5, ROUGE-L, and CLIP-S, respectively. For all tasks, we report the average test performance and standard deviation over ten independent runs.

\subsection{Performance Comparison \textit{(EQ1)}}
\label{subsec:eq1}

To answer Q1, we compare STAGE with a range of competitive baselines. As summarized in Table ~\ref{table:overall performance}, STAGE consistently outperforms all baseline groups, including advanced FGL methods (FedLap, S2FGL, FedSPA, FedIIH), multimodal baselines (Fed-MGNet, Fed-MHGAT, FedMVP, FedMAC), and generic FL / prototype methods (FedAvg, FedProto). Compared with the best advanced FGL baselines, STAGE improves by up to 1.21\% in node classification, 3.60\% in link prediction, and 1.40\% on modality-centric tasks. Against multimodal baselines, the gains reach 1.32\%, 2.52\%, and 4.62\%, respectively. It also surpasses generic FL and prototype methods by up to 4.24\%, 3.74\%, and 2.20\%. These results show that STAGE provides more robust semantic calibration and propagation control across both graph-centric and modality-centric settings.

\begin{table*}[!t]
    \centering
    \caption{Overall performance comparison (mean $\pm$ std, \%). Best results are in \colorbox[HTML]{DADADA}{\textbf{bold}} and second-best are \underline{underlined}.}
    \vspace{-1pt}
    
    \setlength{\tabcolsep}{4pt} 
    \resizebox{\textwidth}{!}{
    \begin{tabular}{c|c|c c | c c | c c | c | c}
    \specialrule{1.5pt}{1.5pt}{1.5pt}

    \multicolumn{2}{c|}{{\textbf{Description}}} 
    & \multicolumn{2}{c|}{\begin{tabular}{@{}c@{}}\textbf{Node Classification} \\ \textbf{(Acc)}\end{tabular}} 
    & \multicolumn{2}{c|}{\begin{tabular}{@{}c@{}}\textbf{Link Prediction} \\ \textbf{(AUC)}\end{tabular}} 
    & \multicolumn{2}{c|}{\begin{tabular}{@{}c@{}}\textbf{Modal Retrieval} \\ \textbf{(R@5)}\end{tabular}} 
    & \begin{tabular}{@{}c@{}}\textbf{G2Text} \\ \textbf{(ROUGE-L)}\end{tabular} 
    & \begin{tabular}{@{}c@{}}\textbf{G2Image} \\ \textbf{(CLIP-S)}\end{tabular} \\ 
    \cmidrule(lr){1-2} \cmidrule(lr){3-4} \cmidrule(lr){5-6} \cmidrule(lr){7-8} \cmidrule(lr){9-9} \cmidrule(lr){10-10}
    
    \multicolumn{2}{c|}{\textbf{Subgraph-FL}} & \textbf{Toys} & \textbf{Grocery} & \textbf{Bili Music} & \textbf{DY} & \textbf{QB} & \textbf{Bili Cartoon} & \textbf{Flickr30k} & \textbf{SemArt} \\ 
    \midrule

    % FL
    \cellcolor[HTML]{B5C3D7} FL & \cellcolor[HTML]{B5C3D7} FedAvg & 
    $75.76_{\scriptstyle \pm 0.20}$ & 
    $74.53_{\scriptstyle \pm 0.18}$ & 
    $65.19_{\scriptstyle \pm 1.30}$ & 
    $64.91_{\scriptstyle \pm 1.78}$ & 
    $84.10_{\scriptstyle \pm 6.62}$ & 
    $74.17_{\scriptstyle \pm 2.01}$ &
    $43.75_{\scriptstyle \pm 2.17}$ & 
    $71.31_{\scriptstyle \pm 0.26}$ \\ 
    
    % MM-GNN
    \cellcolor[HTML]{B6C9C0} & \cellcolor[HTML]{B6C9C0} Fed-MGNet  & 
    $67.02_{\scriptstyle \pm 0.56}$ & 
    $71.48_{\scriptstyle \pm 1.02}$ & 
    $66.15_{\scriptstyle \pm 1.32}$ & 
    $\underline{65.18_{\scriptstyle \pm 0.25}}$ & 
    $84.38_{\scriptstyle \pm 3.19}$ & 
    $73.24_{\scriptstyle \pm 2.49}$ &
    \underline{$48.88_{\scriptstyle \pm 0.65}$} & 
    $70.50_{\scriptstyle \pm 0.42}$ \\ 
    \multirow{-2}{*}{\cellcolor[HTML]{B6C9C0} MM-GNN} & \cellcolor[HTML]{B6C9C0} Fed-MHGAT  & 
    $77.29_{\scriptstyle \pm 1.57}$ & 
    $78.06_{\scriptstyle \pm 2.02}$ & 
    $\underline{67.46_{\scriptstyle \pm 2.01}}$ & 
    $63.80_{\scriptstyle \pm 1.26}$ & 
    $87.74_{\scriptstyle \pm 4.89}$ & 
    $71.47_{\scriptstyle \pm 5.21}$ &
    $47.33_{\scriptstyle \pm 1.49}$ & 
    $57.20_{\scriptstyle \pm 1.63}$ \\ 
    
    % MM-FL
    \cellcolor[HTML]{F5D8B7} & \cellcolor[HTML]{F5D8B7} FedMVP  & 
    $78.98_{\scriptstyle \pm 0.07}$ & 
    $74.41_{\scriptstyle \pm 0.43}$ & 
    $67.00_{\scriptstyle \pm 0.47}$ & 
    $64.75_{\scriptstyle \pm 0.73}$ & 
    $87.11_{\scriptstyle \pm 4.97}$ & 
    $70.36_{\scriptstyle \pm 1.32}$ &
    $48.26_{\scriptstyle \pm 0.60}$ & 
    $71.43_{\scriptstyle \pm 0.11}$ \\ 
    \multirow{-2}{*}{\cellcolor[HTML]{F5D8B7} MM-FL} & \cellcolor[HTML]{F5D8B7} FedMAC  & 
    $78.57_{\scriptstyle \pm 0.25}$ & 
    $\underline{80.06_{\scriptstyle \pm 0.61}}$ & 
    $63.30_{\scriptstyle \pm 0.51}$ & 
    $64.51_{\scriptstyle \pm 1.55}$ & 
    $88.65_{\scriptstyle \pm 2.75}$ & 
    $75.15_{\scriptstyle \pm 2.03}$ &
    $47.95_{\scriptstyle \pm 0.89}$ & 
    $70.84_{\scriptstyle \pm 0.61}$ \\ 
    
    % FGL client-drift
    \cellcolor[HTML]{F1D0C6} & \cellcolor[HTML]{F1D0C6} FedLap  & 
    $78.97_{\scriptstyle \pm 0.31}$ & 
    $79.30_{\scriptstyle \pm 0.22}$ & 
    $64.53_{\scriptstyle \pm 0.19}$ & 
    $63.68_{\scriptstyle \pm 0.53}$ & 
    $90.15_{\scriptstyle \pm 1.80}$ & 
    $71.95_{\scriptstyle \pm 2.64}$ &
    $47.71_{\scriptstyle \pm 0.59}$ & 
    $70.36_{\scriptstyle \pm 0.50}$ \\ 
    \multirow{-2}{*}{\cellcolor[HTML]{F1D0C6} \begin{tabular}{@{}c@{}}FGL-drift\end{tabular}} & \cellcolor[HTML]{F1D0C6} S2FGL  & 
    $76.34_{\scriptstyle \pm 0.49}$ & 
    $73.95_{\scriptstyle \pm 0.35}$ & 
    $66.68_{\scriptstyle \pm 0.13}$ & 
    $64.10_{\scriptstyle \pm 0.06}$ & 
    $\underline{91.23_{\scriptstyle \pm 2.17}}$ & 
    $\underline{78.59_{\scriptstyle \pm 5.33}}$ &
    $48.11_{\scriptstyle \pm 1.27}$ & 
    \underline{$71.56_{\scriptstyle \pm 0.18}$} \\ 
    
    % FGL topology-hetero
    \cellcolor[HTML]{E9CDDF} & \cellcolor[HTML]{E9CDDF} FedSPA  & 
    $77.48_{\scriptstyle \pm 0.36}$ & 
    $73.81_{\scriptstyle \pm 1.67}$ & 
    $67.07_{\scriptstyle \pm 0.12}$ & 
    $63.54_{\scriptstyle \pm 0.22}$ & 
    $84.19_{\scriptstyle \pm 1.65}$ & 
    $76.26_{\scriptstyle \pm 3.89}$ &
    $48.38_{\scriptstyle \pm 0.19}$ & 
    $70.01_{\scriptstyle \pm 0.54}$ \\ 
    \multirow{-2}{*}{\cellcolor[HTML]{E9CDDF} \begin{tabular}{@{}c@{}}FGL-hete\end{tabular}} & \cellcolor[HTML]{E9CDDF} FedIIH  & 
    $\underline{79.19_{\scriptstyle \pm 0.77}}$ & 
    $76.45_{\scriptstyle \pm 0.17}$ & 
    $64.87_{\scriptstyle \pm 0.31}$ & 
    $62.13_{\scriptstyle \pm 0.19}$ & 
    $91.05_{\scriptstyle \pm 3.24}$ & 
    $77.44_{\scriptstyle \pm 3.28}$ &
    $46.99_{\scriptstyle \pm 1.18}$ & 
    $70.98_{\scriptstyle \pm 0.55}$ \\ 
    
    % Prototype
    \cellcolor[HTML]{C8BFD9} Prototype & \cellcolor[HTML]{C8BFD9} FedProto  & 
    $71.91_{\scriptstyle \pm 1.76}$ & 
    $79.41_{\scriptstyle \pm 1.17}$ & 
    $53.67_{\scriptstyle \pm 1.80}$ & 
    $57.89_{\scriptstyle \pm 0.60}$ & 
    $89.42_{\scriptstyle \pm 1.04}$ & 
    $78.23_{\scriptstyle \pm 1.05}$ &
    $47.58_{\scriptstyle \pm 0.42}$ & 
    $71.24_{\scriptstyle \pm 0.06}$ \\
    
    \cmidrule(lr){1-10} 
    
    % Ours
    \multicolumn{2}{c|}{\cellcolor[HTML]{E4E3BF} Ours} &
    \cellcolor[HTML]{DADADA}$\textbf{80.30}_{\scriptstyle \pm \textbf{0.35}}$ & 
    \cellcolor[HTML]{DADADA}$\textbf{80.51}_{\scriptstyle \pm \textbf{0.26}}$ & 
    \cellcolor[HTML]{DADADA}$\textbf{68.93}_{\scriptstyle \pm \textbf{0.36}}$ & 
    \cellcolor[HTML]{DADADA}$\textbf{67.70}_{\scriptstyle \pm \textbf{0.24}}$ & 
    \cellcolor[HTML]{DADADA}$\textbf{91.46}_{\scriptstyle \pm \textbf{2.58}}$ & 
    \cellcolor[HTML]{DADADA}$\textbf{79.77}_{\scriptstyle \pm \textbf{1.39}}$ &
    \cellcolor[HTML]{DADADA}$\textbf{49.78}_{\scriptstyle \pm \textbf{0.83}}$ & 
    \cellcolor[HTML]{DADADA}$\textbf{72.69}_{\scriptstyle \pm \textbf{0.53}}$ \\ 

    \specialrule{1.3pt}{2.0pt}{1.0pt}
    \end{tabular}}
    \vspace{-3pt}
    \label{table:overall performance}
\end{table*}

\subsection{Ablation Study \textit{(EQ2)}}

\label{subsec:eq2}
\begin{table*}[htbp]
    \centering
    \caption{Ablation study on Toys, Grocery, Bili Music, DY, QB, Bili Cartoon, Flickr30k, and SemArt datasets. Each variant removes one key boosting mechanism to validate its contribution.}
    \vspace{-0pt}
    \resizebox{\textwidth}{!}{
    \setlength{\tabcolsep}{2.5pt}
    \begin{tabular}{c|cc|cc|cc|c|c}
    \specialrule{1.5pt}{1.5pt}{1.5pt}
    \multirow{2}{*}[-0.5ex]{\textbf{Variant}} &
    \multicolumn{2}{c|}{\textbf{Node Classification (Acc)}} &
    \multicolumn{2}{c|}{\textbf{Link Prediction (AUC)}} &
    \multicolumn{2}{c|}{\textbf{Modal Retrieval(R@5)}} & 
    \multicolumn{1}{c|}{\textbf{G2Text(ROUGE-L)}} & 
    \multicolumn{1}{c}{\textbf{G2Image(CLIP-S)}} \\
    \cmidrule(lr){2-3} \cmidrule(lr){4-5} \cmidrule(lr){6-7} \cmidrule(lr){8-8} \cmidrule(lr){9-9}
    & \textbf{Toys} & \textbf{Grocery} & \textbf{Bili Music} & \textbf{DY} & \textbf{QB} & \textbf{Bili Cartoon} & \textbf{Flickr30k} & \textbf{SemArt} \\
    \midrule

    Full STAGE
    & \cellcolor[HTML]{DADADA}$\textbf{80.30}_{\scriptstyle \pm \textbf{0.15}}$
    & \cellcolor[HTML]{DADADA}$\textbf{80.51}_{\scriptstyle \pm \textbf{0.06}}$
    & \cellcolor[HTML]{DADADA}$\textbf{68.93}_{\scriptstyle \pm \textbf{0.36}}$
    & \cellcolor[HTML]{DADADA}$\textbf{67.70}_{\scriptstyle \pm \textbf{0.24}}$ 
    & \cellcolor[HTML]{DADADA}$\textbf{80.32}_{\scriptstyle \pm \textbf{0.39}}$
    & \cellcolor[HTML]{DADADA}$\textbf{77.79}_{\scriptstyle \pm \textbf{1.49}}$ 
    & \cellcolor[HTML]{DADADA}$\textbf{50.14}_{\scriptstyle \pm \textbf{0.85}}$ 
    & \cellcolor[HTML]{DADADA}$\textbf{72.69}_{\scriptstyle \pm \textbf{0.53}}$ \\

    w/o Frozen Semantic Bank
    & $76.54_{\scriptstyle \pm 0.21}$
    & $75.87_{\scriptstyle \pm 0.16}$
    & $65.72_{\scriptstyle \pm 0.12}$
    & $64.92_{\scriptstyle \pm 0.32}$ 
    & $75.86_{\scriptstyle \pm 1.20}$
    & $74.71_{\scriptstyle \pm 1.69}$ 
    & $46.80_{\scriptstyle \pm 0.81}$ 
    & $67.28_{\scriptstyle \pm 0.54}$ \\

    w/o Contrastive GAP
    & $78.69_{\scriptstyle \pm 0.44}$
    & $76.82_{\scriptstyle \pm 0.28}$
    & $67.73_{\scriptstyle \pm 0.45}$
    & $65.85_{\scriptstyle \pm 0.16}$ 
    & $75.77_{\scriptstyle \pm 1.71}$
    & $73.77_{\scriptstyle \pm 0.96}$ 
    & $45.05_{\scriptstyle \pm 0.88}$ 
    & $65.25_{\scriptstyle \pm 0.31}$ \\

    w/o Meta-Controller
    & \underline{$79.29_{\scriptstyle \pm 0.12}$}
    & $79.91_{\scriptstyle \pm 0.17}$
    & $67.80_{\scriptstyle \pm 0.20}$
    & $66.84_{\scriptstyle \pm 0.33}$ 
    & \underline{$76.27_{\scriptstyle \pm 0.52}$}
    & \underline{$77.51_{\scriptstyle \pm 1.71}$} 
    & \underline{$47.87_{\scriptstyle \pm 0.96}$} 
    & $69.59_{\scriptstyle \pm 0.39}$ \\

    w/o Max-Entropy Projection
    & $79.18_{\scriptstyle \pm 0.51}$
    & \underline{$80.34_{\scriptstyle \pm 0.15}$}
    & \underline{$68.01_{\scriptstyle \pm 0.57}$}
    & \underline{$66.97_{\scriptstyle \pm 0.53}$} 
    & $75.37_{\scriptstyle \pm 0.89}$
    & $75.16_{\scriptstyle \pm 1.82}$ 
    & $46.02_{\scriptstyle \pm 0.98}$ 
    & \underline{$71.28_{\scriptstyle \pm 0.26}$} \\

    \specialrule{1.3pt}{2.0pt}{1.0pt}
    \end{tabular}}
    \vspace{-3pt}
    \label{table:ablation}
\end{table*}

To address EQ2, we analyze the key components of the STAGE framework.

% The frozen anchor bank and the GAP resolve the root problem of feature drift, while the differentiable homophily controller and entropy regularization ($\mathcal{L}_{\mathrm{ent}}^{(k)}$) are applied to suppress downstream consequences. 

\textbf{Module Ablation.}
The ablation results in Table~\ref{table:ablation} show that removing any key module degrades performance. In particular, excluding the Frozen Semantic Bank causes the largest average drop (3.84\%), with especially severe degradation on SemArt (5.41\%), confirming the necessity of a shared semantic space for multimodal collaboration. Removing Contrastive GAP also leads to a large average decline (3.68\%), highlighting its role in resolving feature drift before graph propagation. Without Max-Entropy Projection, performance drops notably on modality retrieval for QB (4.95\%) and G2Text on Flickr30k (4.12\%), indicating its importance in preventing protocol collapse. Finally, disabling the meta-controller yields a consistent average drop of 1.66\%, showing that propagation regulation remains necessary for suppressing downstream inconsistency amplification. Overall, all components contribute meaningfully to the effectiveness of STAGE.

\subsection{Robustness Analysis \textit{(EQ3)}}
\label{subsec:eq3}

To answer EQ3, we evaluate STAGE under more hostile data conditions and across a range of key hyperparameters.

\begin{figure*}[htbp]
    \centering
    \includegraphics[width=\textwidth]{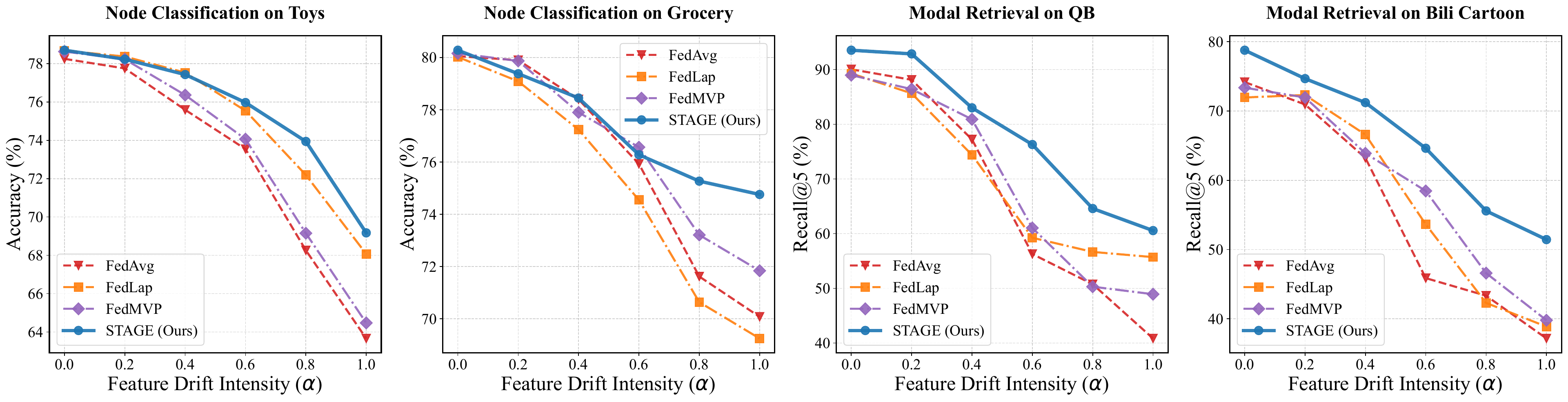}
    \caption{\textbf{Robustness against feature drift.} Performance decay of different methods across four datasets as feature drift intensity ($\alpha$) increases from 0.0 to 1.0. STAGE exhibits superior resilience, whereas parameter-centric and uncalibrated multimodal baselines suffer from severe degradation.}
    \label{fig:feature_drift_robustness}
\end{figure*}

\textbf{Robustness to Feature Drift.} To control cross-client semantic mismatch, we introduce a feature Non-IID intensity parameter $\alpha \in [0.0, 1.0]$. Specifically, for each client $k$, we shift its local feature space by a deterministic Gaussian bias $\alpha \cdot \mathbf{v}_k$, where $\mathbf{v}_k \sim \mathcal{N}(\mathbf{0}, \mathbf{I})$. Larger $\alpha$ induces stronger centroid drift and thus more challenging MM-FGL conditions. As shown in Fig.~\ref{fig:feature_drift_robustness}, when $\alpha$ is small, most baselines remain competitive. However, as $\alpha$ approaches 1.0, parameter-centric FL (FedAvg), topology-aware FGL (FedLap), and uncalibrated multimodal FL (FedMVP) degrade sharply. In contrast, STAGE exhibits a much slower decay across all four datasets, showing that its GAP-based calibration effectively preserve cross-client semantic comparability under severe feature drift.

\textbf{Robustness to Modality Corruption.} 
\begin{figure*}[htbp]
    \centering
    \includegraphics[width=\textwidth]{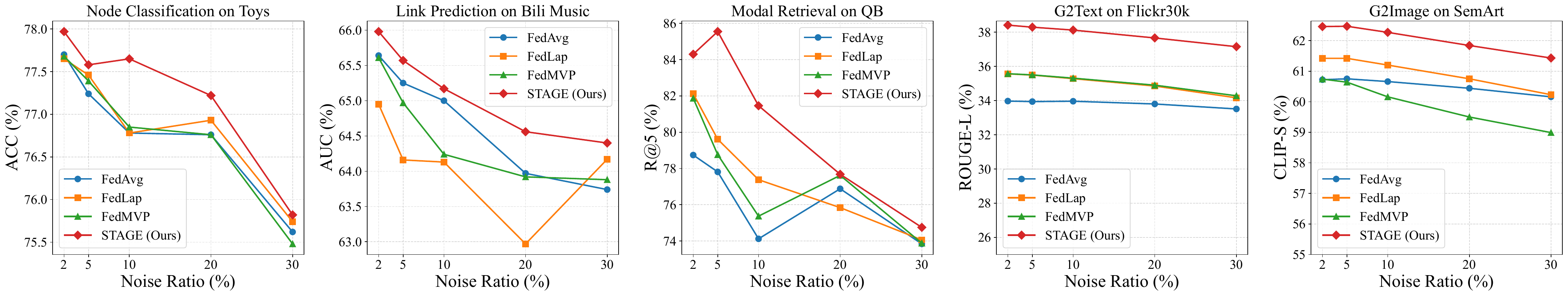}
    \caption{Robustness against modality noise on QB, Bili Music, Toys, Flickr30k, and SemArt datasets. We randomly replace the input text and image embeddings with noise at varying ratios ($2\%$ to $30\%$). STAGE exhibits superior resilience compared to other federated graph learning baselines.}
    \label{fig:noise_robustness}
\end{figure*}
To evaluate robustness to noisy modalities, we replace textual and visual embeddings with Gaussian noise at ratios from $2\%$ to $30\%$. As shown in Fig.~\ref{fig:noise_robustness}, most baselines remain competitive under mild noise but degrade rapidly as corruption increases. In contrast, STAGE exhibits a much slower and more stable decay across all datasets. This shows that, by projecting local features into a globally calibrated anchor simplex and adaptively down-weighting unreliable neighborhoods, STAGE can better contain modality noise and reduce its amplification during graph propagation.

\begin{figure}[htbp]
  \centering
  \includegraphics[width=0.48\columnwidth]{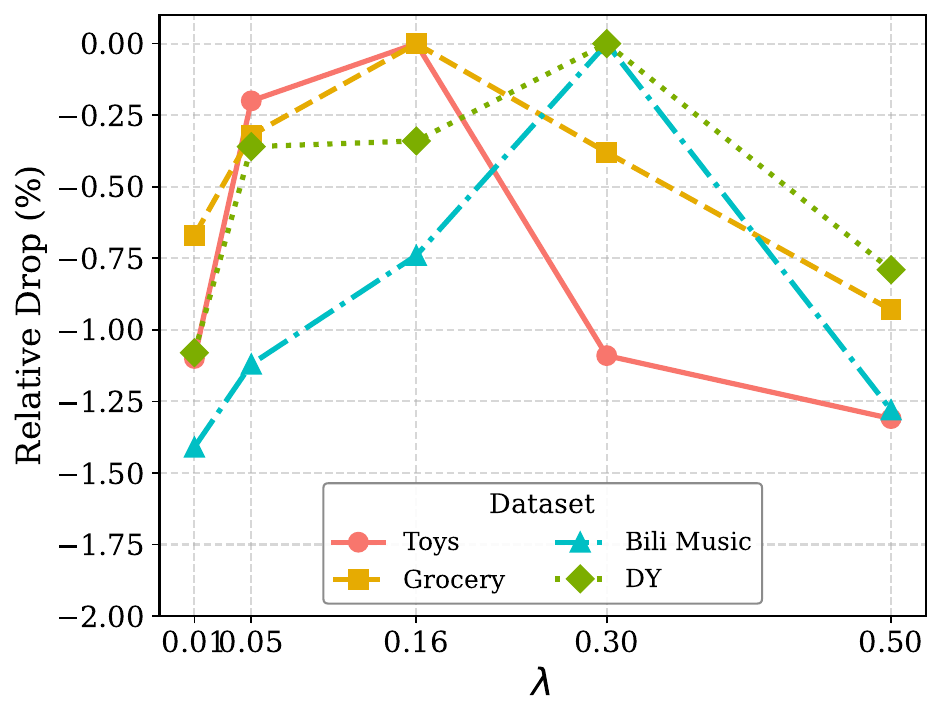}
  \hfil
  \includegraphics[width=0.48\columnwidth]{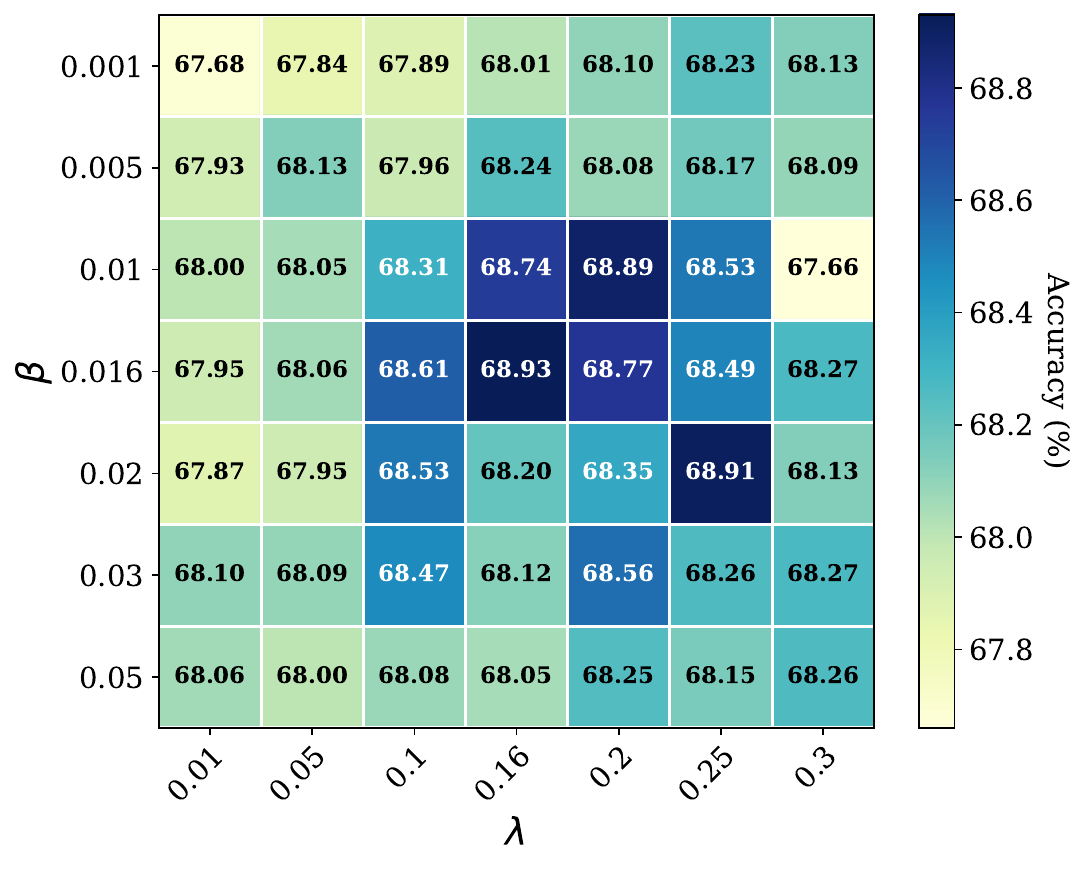}
  \caption{\textbf{Hyperparameter sensitivity.} (Left) Relative performance drop under different semantic calibration weights $\lambda$. (Right) Heatmap over semantic calibration and entropy regularization hyperparameters ($\lambda, \beta$).}
  \label{fig:hyperparameter}
\end{figure}

\textbf{Hyperparameter Sensitivity.}We perform a sensitivity analysis on key hyperparameters in STAGE. For the InfoNCE semantic calibration ($\mathcal{L}_{gap}$), which resolves feature drift, we vary its strength coefficient $\lambda$ (Eq. 13). For the Max-Entropy projection ($\mathcal{L}_{ent}$), designed to prevent protocol collapse, we vary its intensity regulator $\beta$ (Eq. 13). Results are shown in Fig.~\ref{fig:hyperparameter}: (Left) varying $\lambda$ exhibits a parabolic trend with a broad, high-performing plateau (optimal near $\lambda \in [0.16, 0.30]$) across four diverse datasets, indicating stable semantic calibration; (Right) the joint sensitivity heatmap reveals a broad and continuous dark-blue high-performance region for combinations of $\lambda$ and $\beta$. Overall, STAGE shows strong robustness to key hyperparameters and the interplay between semantic calibration and entropy regularization.

\begin{figure}[!t]
  \centering
  \includegraphics[width=0.8\columnwidth]{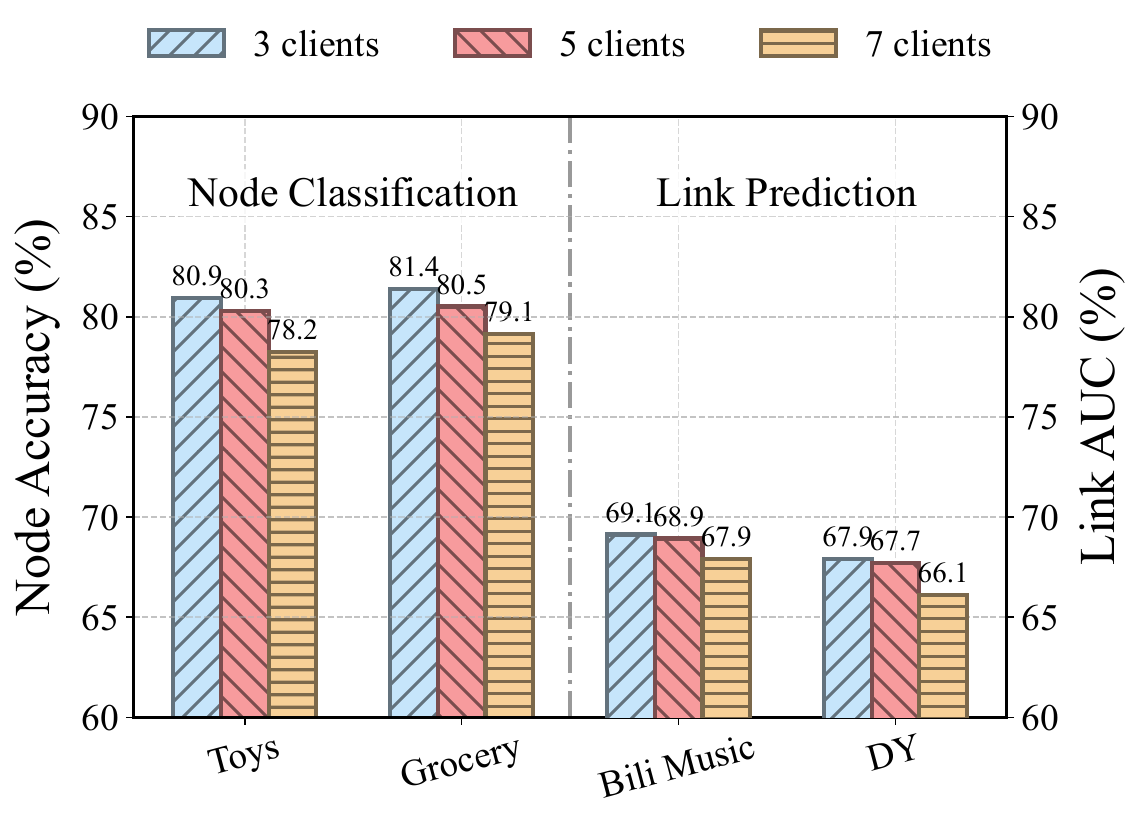}
  \caption{\textbf{Scalability across Client Partitions.} Performance impact of scaling the number of decentralized clients ($K \in \{3, 5, 7\}$).}
  \vspace{-6mm}
  \label{fig:client effect}
\end{figure}

\textbf{Scalability to Client Fragmentation.} Fig.~\ref{fig:client effect} further evaluates STAGE under increasing client fragmentation. As the number of clients grows, each local subgraph becomes smaller and less structurally complete, making cross-client collaboration more difficult. Nevertheless, STAGE maintains relatively stable performance on both node classification and link prediction. Although a mild decline is observed when the client number increases from 3 to 7, the overall drop remains limited, indicating that STAGE is robust to highly fragmented subgraphs and can preserve effective cross-client semantic coordination under stronger partitioning.

\subsection{Communication Efficiency \textit{(EQ4)}}
\label{subsec:eq4}

To answer EQ4, we compare the per-round communication payload and computational complexity of STAGE against representative baselines in Table~\ref{tab:efficiency_final}.

\begin{table}[htbp]
    \centering
    \caption{\textbf{Efficiency and resource comparison} }
    \label{tab:efficiency_final}
    \renewcommand{\arraystretch}{1.2}
    \setlength{\tabcolsep}{3pt}
    \resizebox{\columnwidth}{!}{
    \begin{tabular}{lcccc}
        \toprule
        \textbf{Method} & \textbf{Comm. Payload} & \textbf{Total Ops} & \textbf{Space Cost} & \textbf{Time} \\
        & (Scalars) & (FLOPs) & (MB) & (s) \\
        \midrule
        FedAvg & $1.00 \times 10^6$ & \boldmath$1.7 \times 10^8$ & $44.9$ & \textbf{0.0891} \\
        FedMAC & $1.28 \times 10^5$ & $3.1 \times 10^8$ & $41.2$ & $0.1599$ \\
        FedSPA & $1.00 \times 10^6$ & $3.6 \times 10^9$ & $45.8$ & $1.8257$ \\
        \rowcolor[gray]{0.9} STAGE (Ours) & \boldmath$8.19 \times 10^3$ & $7.3 \times 10^8$ & \textbf{6.2} & 0.4225 \\
        \bottomrule
    \end{tabular}
    }
\end{table}
\textbf{Communication and Memory Footprint.}
Table~\ref{tab:efficiency_final} shows that STAGE achieves high efficiency by restricting inter-client coordination to a low-dimensional protocol space. Under our setup, this yields a per-round communication payload of only 8,192 scalars ($M=128$, $d_p=64$), a 122$\times$ reduction compared with FedAvg and FedSPA, while local memory usage drops to 6.2 MB, about 7$\times$ smaller than the baselines. Although STAGE incurs modest extra computation ($7.3\times10^8$ FLOPs, 0.4225s) over standard FL due to semantic calibration and propagation control, it remains substantially more efficient than graph-specific baselines such as FedSPA. This overhead is moderate and necessary for mitigating feature drift.

\begin{figure}[htbp]
  \centering
  \vspace{0.3cm} 
  \includegraphics[width=0.48\columnwidth]{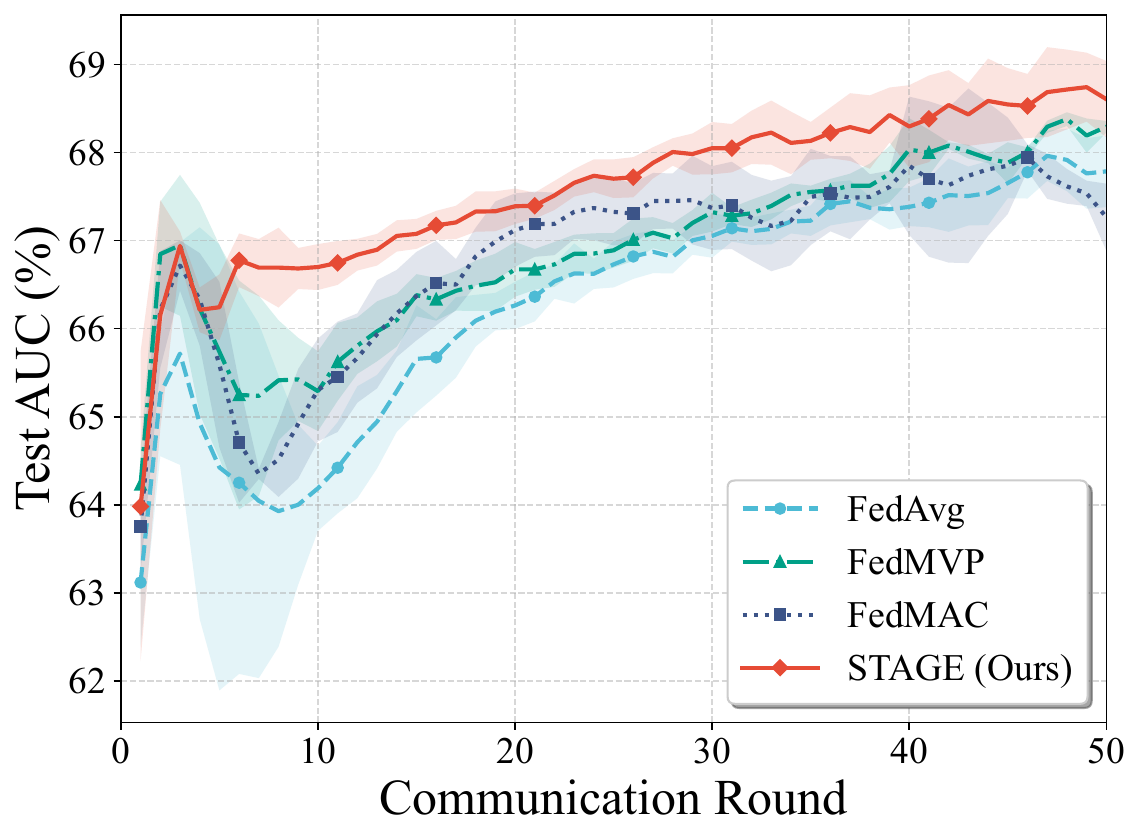}
  \hfil
  \includegraphics[width=0.48\columnwidth]{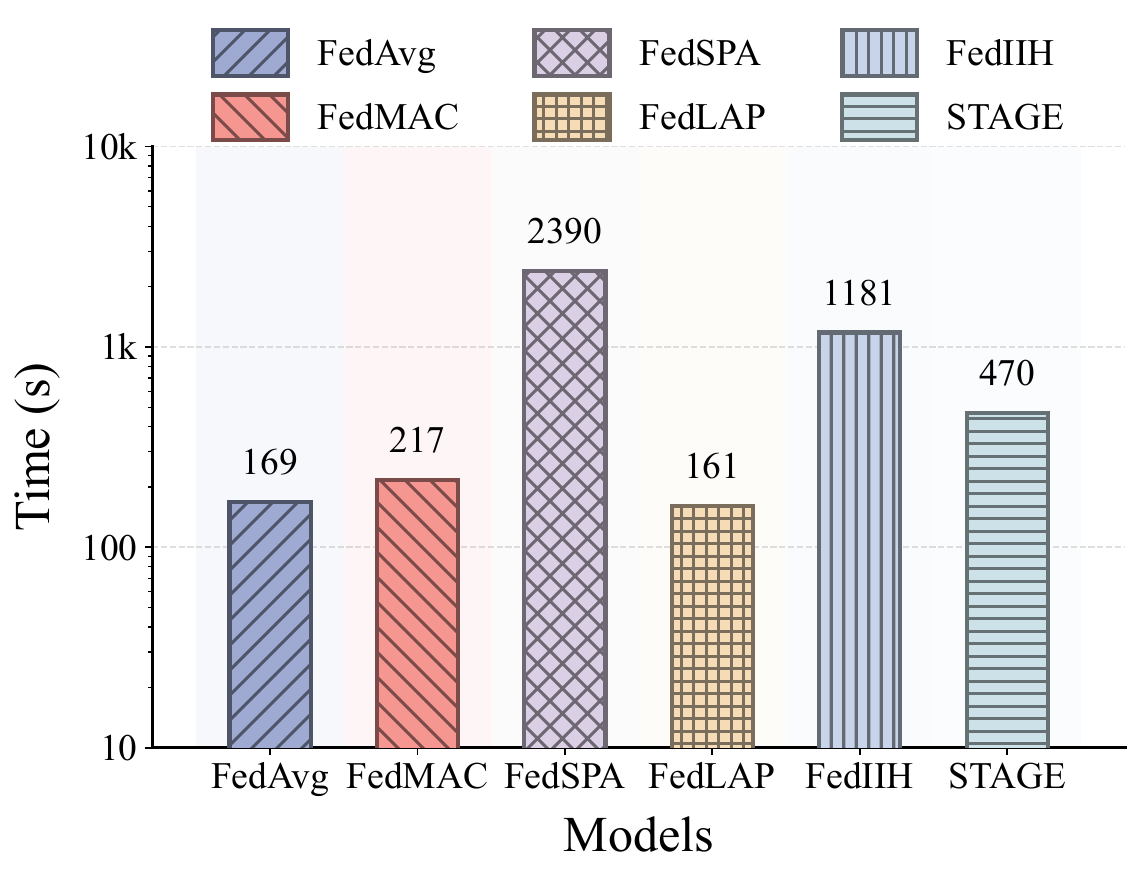}
  \caption{\textbf{Efficiency Analysis on Bili Music.} (Left) STAGE achieves faster convergence and a higher AUC ceiling compared to baselines. (Right) Total computational time demonstrates that STAGE adds only marginal overhead to FedAvg while being significantly faster than complex FGL methods.}
  \label{fig:overall_efficiency}
  \vspace{0.2cm} 
\end{figure}

\textbf{Accelerated Training Convergence.} Fig.~\ref{fig:overall_efficiency} shows that STAGE converges faster and reaches a higher AUC ceiling on Bili Music, indicating more stable decentralized training. Meanwhile, its computational cost remains in the same order as standard FL, with only modest extra overhead from semantic calibration and meta-controller inference, which are necessary to resolve the feature drift problem.

\section{Conclusion}
\label{sec:conclusion}
In this paper, we presented \textbf{STAGE}, a protocol-first framework for multimodal federated graph learning. Our work is motivated by a clear failure hierarchy in MM-FGL: \textbf{feature drift} is the root problem, while \emph{pseudo-alignment} and \emph{propagation-induced drift} arise as downstream failures when cross-client semantic inconsistency is not resolved before graph collaboration. To address this challenge, STAGE separates the learning process into two ordered stages. First, \textbf{Variational Semantic Calibration} maps heterogeneous multimodal features into a shared anchor space and calibrates anchor meanings across clients through server-side prototypes, thereby improving semantic comparability and mitigating pseudo-alignment. Second, \textbf{Differentiable Homophily Control} regulates graph propagation to suppress the amplification of residual inconsistency over local neighborhoods.
Our theoretical analysis supports this design through a generalization guarantee for semantic calibration and a contraction guarantee for regulated propagation. Extensive experiments further show that STAGE consistently achieves strong performance across multiple multimodal graph tasks, remains robust under severe feature drift and client fragmentation, and reduces communication cost through low-dimensional protocol messages. Overall, we hope this work highlights the importance of resolving semantic inconsistency before aggregation and propagation, and encourages future research on protocol-level designs for decentralized multimodal graph learning.

\bibliographystyle{IEEEtran}
\bibliography{reference}

\vspace{-2cm}
\begin{IEEEbiography}
[{\includegraphics[width=0.8in,height=1in,clip,keepaspectratio]{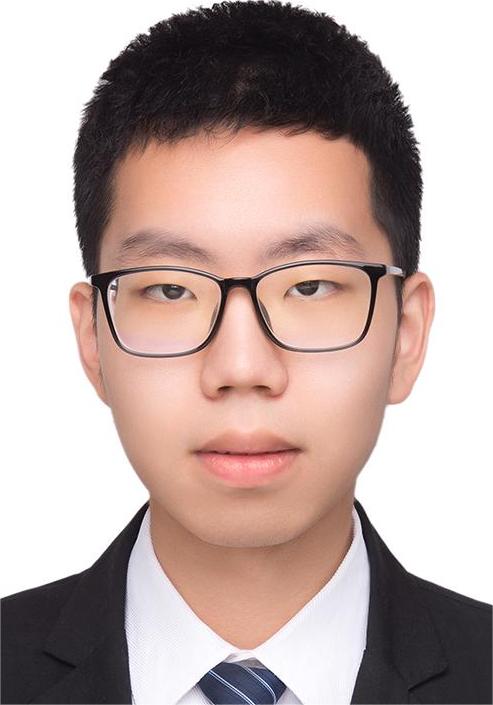}}]{Zekai Chen}
is currently pursuing his Master's degree in Computer Science at Beijing Institute of Technology under the supervision of Professor Rong-Hua Li. He obtained his Bachelor degree in Computer Science from the same institution in 2024. His research focuses on Graph Machine Learning and AI for Science (AI4Science). 
\end{IEEEbiography}
\vspace{-2.5cm}
\begin{IEEEbiography}
[{\includegraphics[width=0.8in,height=1in,clip,keepaspectratio]{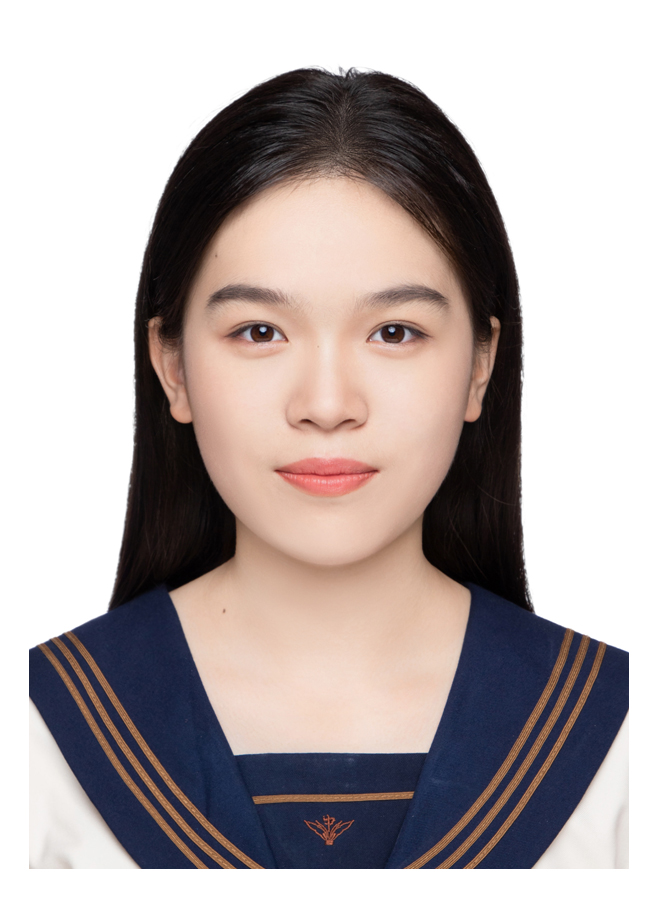}}]{Xun Wu} is currently pursuing the B.S. degree in data science and big data technology with the School of Computer Science, Beijing Institute of Technology, Beijing, China. Her research interest lies in Graph Machine Learning.
\end{IEEEbiography}
\vspace{-2.5cm}
\begin{IEEEbiography}[{\includegraphics[width=0.8in,height=1in,clip,keepaspectratio]{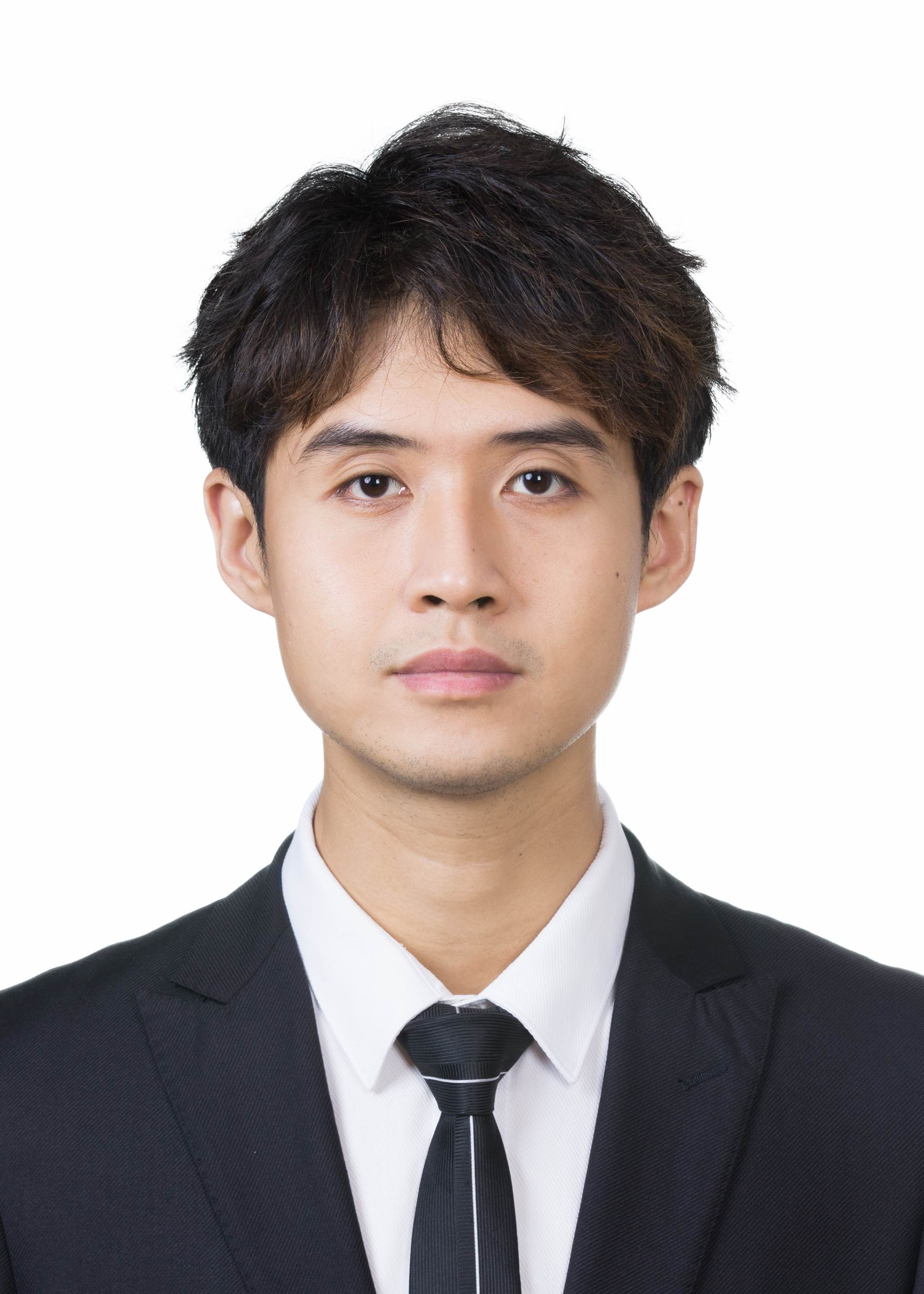}}]{Xunkai Li}
is currently pursuing the PhD degree in Beijing Institute of Technology, advised by Prof. Rong-Hua Li. He received the BS degree from Shandong University in 2022. His research interest lies in Data-centric Graph Intelligence (Data-centric AI, Graph Machine Learning, and AI4Science). He has published 10+ papers in top ML/DB/DM/AI conferences such as ICML, VLDB, WWW, AAAI.
\end{IEEEbiography}
\vspace{-2.5cm}
\begin{IEEEbiography}
[{\includegraphics[width=0.8in,height=1in,clip,keepaspectratio]{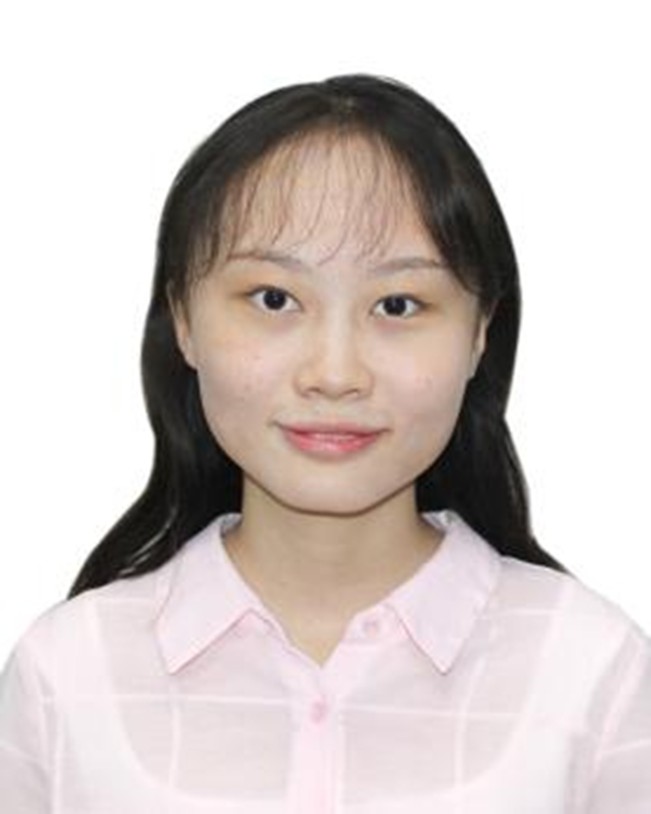}}]{Yihan Sun}
Yihan Sun, was born in Beijing, China, 2006. She is currently a second-year undergraduate student majoring in Computer Science and Technology at Minzu University of China, Beijing, China. Her research interests include computer science and technology.
\end{IEEEbiography}
\vspace{-2.5cm}
\begin{IEEEbiography}[{\includegraphics[width=0.8in,height=1in,clip,keepaspectratio]{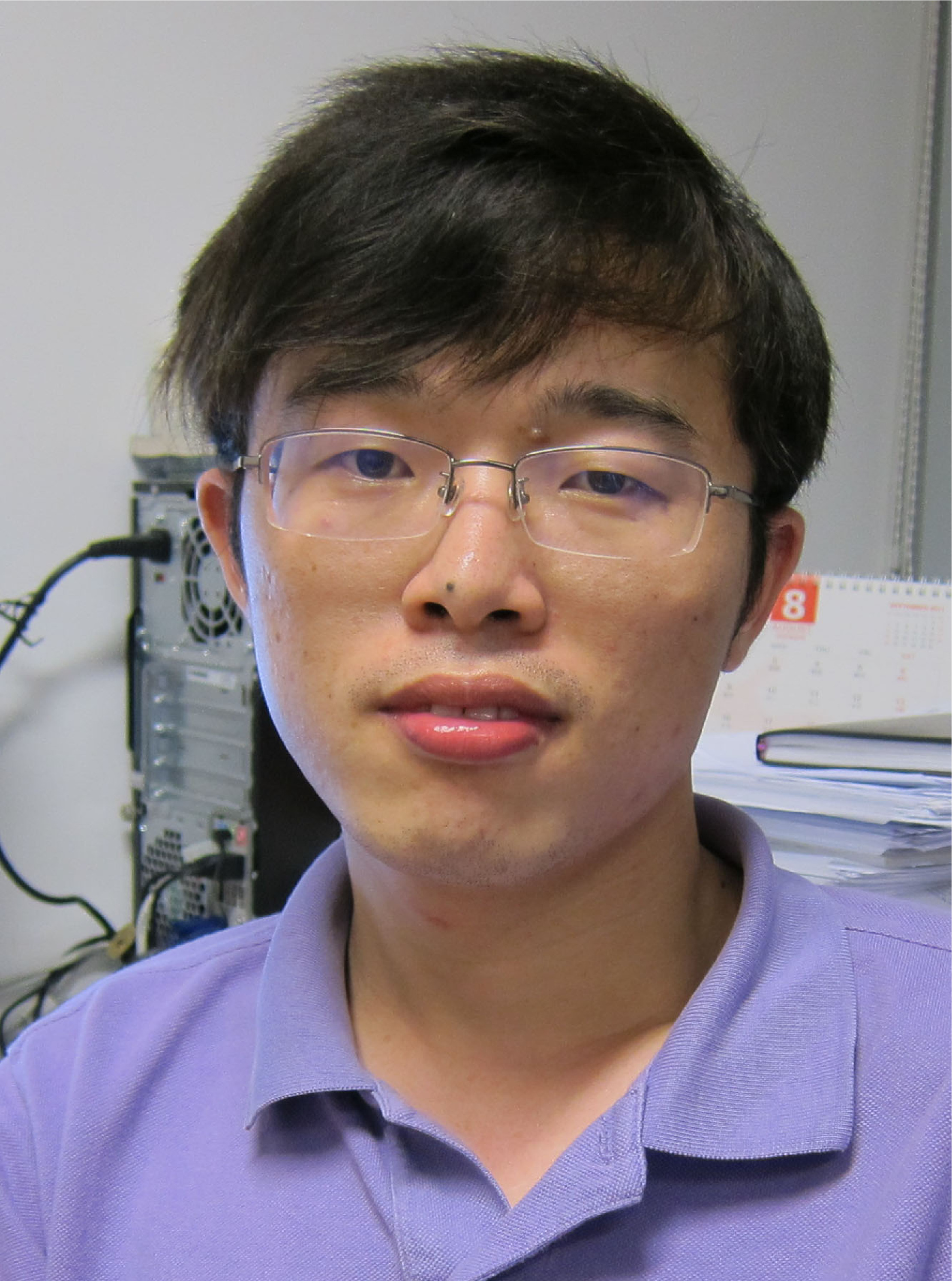}}]{Rong-Hua Li}
 received the PhD degree from the Chinese University of Hong Kong, in 2013. He is currently a professor with the Beijing Institute of Technology (BIT), Beijing, China. His research interests include graph data management and mining, social network analysis, graph computational systems, and graph-based machine learning.
\end{IEEEbiography}
\vspace{-2.5cm}
\begin{IEEEbiography}[{\includegraphics[width=0.8in,height=1in,clip,keepaspectratio]{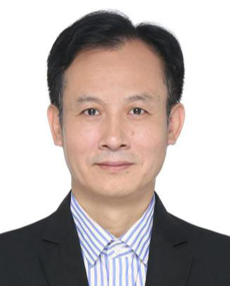}}]{Guoren Wang}
 received the BS, MS, and PhD degrees from the Department of Computer Science, Northeastern University, China, in 1988, 1991, and 1996, respectively. Currently, he is a professor with the Beijing Institute of Technology (BIT), Beijing, China. His research interests include graph data management, graph mining, and graph computational systems.
\end{IEEEbiography}
\vspace{-2.5cm}
\end{document}